# Automatic Construction of Multiple Classification Dimensions for Managing Approaches in Scientific Papers

Bing Ma and Hai Zhuge

**Abstract**—Approaches form the foundation for conducting scientific research. Querying approaches from a vast body of scientific papers is extremely time-consuming, and without a well-organized management framework, researchers may face significant challenges in querying and utilizing relevant approaches. Constructing multiple dimensions on approaches and managing them from these dimensions can provide an efficient solution. Firstly, this paper identifies approach patterns using a top-down way, refining the patterns through four distinct linguistic levels: semantic level, discourse level, syntactic level, and lexical level. Approaches in scientific papers are extracted based on approach patterns. Additionally, five dimensions for categorizing approaches are identified using these patterns. This paper proposes using tree structure to represent step and measuring the similarity between different steps with a tree-structure-based similarity measure that focuses on syntactic-level similarities. A collection similarity measure is proposed to compute the similarity between approaches. A bottom-up clustering algorithm is proposed to construct class trees for approach components within each dimension by merging each approach component or class with its most similar approach component or class in each iteration. The class labels generated during the clustering process indicate the common semantics of the step components within the approach components in each class and are used to manage the approaches within the class. The class trees of the five dimensions collectively form a multi-dimensional approach space. The application of approach queries on the multi-dimensional approach space demonstrates that querying within this space ensures strong relevance between user queries and results and rapidly reduces search space through a class-based query mechanism.

**Index Terms**—Approach Patterns, Extract Complete Approaches, Similarity Calculation, Class Tree, Multiple Dimensions, Approach Query

## 1 INTRODUCTION

APPROACHES form the foundation of scientific research. In scientific papers, the approaches describe detailed procedures employed during the study, enabling readers to clearly understand how the research was conducted. Efficient management of approaches is critical for improving the efficiency of approach retrieval and utilization. Searching for specific appraoches from a vast number of scientific papers can be highly time-consuming, and without a structured management framework, researchers may face significant challenges in querying and utilizing relevant approaches.

An efficient way to managing a large number of approaches is to establish classification dimensions. A classification dimension refers to a specific perspective from which approaches can be categorized. Different dimensions reflect distinct perspectives and criteria for classifying approaches, with multiple dimensions collectively forming a multi-dimensional classification space [1][2]. Within this space, researchers can efficiently locate and access relevant appraoch classes by navigating along specific dimensions according to their needs.

Before constructing dimensions on approaches, a challenging problem is extracting approaches from scientific papers. Previous studies on approach extraction have primarily concentrated on the extraction of approach sentences or approach entities (i.e., words or phrases) [3][4]. Single sentence or entity conveys limited information and often lacks the necessary depth and context required for a comprehensive understanding of the intricate approaches employed. Actually, approaches in scientific papers are typically presented as sequences of sentences that outline the process for addressing specific problems or achieving particular goals. A complete approach usually consists of a series of steps, each described by one or more sentences. For example, an approach in the paper "Automatic text summarization with neural networks" is "*A new technique for summarizing news articles using a neural network is presented. A neural network is trained to learn the relevant characteristics of sentences that should be included in the summary of the article. The neural network is then modified to generalize and combine the relevant characteristics apparent in summary sentences. Finally, the modified neural network is used as a filter to summarize news articles.*", which consists of three steps, with each step expressed by a complete sentence.

Therefore, to enhance the comprehensibility and usability of approaches, this study aims to extract complete approaches comprising multiple steps. Specifically, we identify approach patterns using a top-down way, refining them through four distinct linguistic levels: semantic

• *Bing Ma is with the Key Lab of Intelligent Information Processing at the Institute of Computing Technology in the Chinese Academy of Sciences and the School of Computer and Control Engineering in the University of Chinese Academy of Sciences. E-mail: mabing17z@ict.ac.cn*
• ** Corresponding author: Hai Zhuge is a chair professor of the Great Bay University. E-mail: zhuge@gbu.edu.cn*

level, discourse level, syntactic level, and lexical level. This gradual refinement of approach patterns intends to achieve more comprehensive extraction of approaches from scientific papers and enhance the scalability of patterns.

Based on the approach patterns, five distinct dimensions for categorizing approaches can be derived, and the approach components within each dimension can be identified using these patterns. Specifically, each approach component within one dimension consists of a collection of step components, where a step component can be a sentence, a part of a sentence (such as a phrase or a word), or even multiple sentences. Considering the language characteristics of step components, this paper proposes a tree-structure-based similarity measure to calculate the similarity between step components, which focuses on syntactic-level similarities. Furthermore, a novel collection similarity measure is proposed to compute the similarity between approach components, as each approach component is a collection of step components. The combination of these two similarity measures enables approach components with step components that share similar major syntactic constituents receiving a higher similarity score compared to those with step components sharing similar minor syntactic constituents.

Furthermore, a bottom-up clustering algorithm is proposed to construct class trees for approach components within each dimension by merging each approach component or class with its most similar approach component or class in each iteration based on the proposed similarity measures. The common elements shared by the step components of the approach components in each class are used as class labels to manage the approaches within that class. The clustering process allows an approach component to belong to multiple classes, as it may share different common step components with approach components of different classes.

The collections of class trees constructed from the approach components within each dimension constitutes that dimension. Five distinct dimensions form the multi-dimensional approach space, providing an efficient means for managing approaches. To verify the effectiveness of using the multi-dimensional approach space for managing approaches, this paper discusses the application of approach queries on the multi-dimensional approach space. This demonstrates that querying within the multi-dimensional approach space ensures strong relevance between user queries and results and rapidly reduces the search space through a class-based query mechanism.

## 2 Approach Patterns Within Scientific Papers

Approaches in scientific papers typically refer to a structured procedure or process that comprises a sequential series of steps, executed in a prescribed order. Some approaches may also incorporate a general description preceding or following the sequential steps, serving to summarize or generalize the overall process.

Generally, approaches are proposed for specific purposes, requiring the execution of a series of steps to gradually achieve those purposes. Each step comprises an action or operation, which needs to be executed using specific manners or techniques and performed under certain conditions. Upon completion of the actions, certain effects or results are achieved.

The high-level pattern of approach can be generalized as follows:

```
<Approach> ::= [<general description>] <step>
      {<step>} [<general description>]
```

The pattern illustrates that the approaches we focus on are those that contain at least one step. Each step comprises one or multiple sentences, where certain syntactic roles (e.g., subject, verb, object, or adverbial) conform to specific patterns (e.g., containing specific identifiers). Similarly, each general description also comprises one or multiple sentences, with certain syntactic roles adhering to specific patterns that identify general descriptions.

### 2.1 General Idea for Identifying Approach Patterns

This paper identifies approach patterns using a top-down way, which gradually refines the patterns through four distinct levels: semantic level, discourse level, syntactic level, and lexical level.

Steps form the core of an approach. In scientific papers, a single step of an approach may contain one or multiple sentences, necessitating an analysis that transcends the syntactic category and extends into the realm of discourse analysis. Discourse refers to a linguistic unit that is larger and more complex than a single sentence, encompassing a comprehensive assemblage of sentences that collectively convey a coherent and meaningful narrative [7].

In linguistics, discourse relations refer to the logical or semantic connections between sentences or sentence elements within a given discourse [8]. Inspired by the PDTB (Penn Discourse TreeBank) sense hierarchy[?], this paper proposes five discourse relations that can be utilized to express steps and sequential processes:

Sequential relation emphasizes the temporal order or the sequence of steps within an approach process. It is particularly significant when describing a series of consecutive steps. Therefore, sequential relation stands as an important type of discourse relation for expressing steps in approaches.

Purpose-action relation is a type of discourse relation, which denotes a connection where one sentence or sentence element describes a purpose, while another sentence or sentence element details an action taken to achieve that purpose. This relation is important in delineating a step, as it is often necessary to clarify, within a step, not only what action or operation is being taken but also the purpose behind that action or operation.

Action-manner relation, another significant type of discourse relation, plays a pivotal role in describing steps. It represents a connection where one sentence or sentence element describes the manner or adopted technique in which the action or operation, detailed in another sentence or sentence element, has been performed.

Cause-effect relation is a type of discourse relation,

where one sentence or sentence element describes the cause and another describes the result or effect caused by the former. This relation is equally applicable in describing a step. In one scenario, the action or operation of a step can correspond to the result or effect, with the specific manner or technique used in executing that action or operation corresponding to the cause. In another scenario, the action or operation can correspond to the cause, and if it is executed, it can lead to the result or effect described in another sentence or sentence element.

Condition-consequence relation is another type of discourse relation, where one sentence or sentence element describes a situation as unrealized, and when it is realized, it would lead to the consequence described by another sentence or sentence element. This relation can also be utilized to delineate a step. The action or operation of a step can be considered as a consequence, meaning that if a specified condition described in one sentence or sentence element is met, the action or operation can be successfully executed. One scenario is where the specific manner or technique employed serves as a condition for executing the action or operation.

Other discourse relations in the PDTB sense hierarchy, such as relations under the COMPARISON class and relations under the EXPANSION class other than Manner relation, are typically not used to express steps in scientific papers, as these relations are generally not employed to describe actions or manners [9].

At the semantic level, step patterns are refined by categorizing them based on five discourse semantic relations: sequential relation, purpose-action relation, action-manner relation, cause-effect relation, and condition-consequence relation. Figure 1 visually demonstrates the gradual refinement process of the patterns, using the step with purpose-action relation as an example to illustrate the further refinement process after the semantic level. Directly refining and identifying the step patterns of each semantic relation through lexical items that represent the relation becomes challenging due to the difficulty in enumerating all possible lexical items. As there are numerous words or phrases that can convey the same meaning, and the lexical items expressing the same semantic relation may contain diverse parts of speech or their varied combinations. Therefore, this paper gradually refines the step patterns of each semantic relation from the discourse level, syntactic level, and lexical level.

A discourse typically comprises one or multiple sentences, with syntax serving as the foundation for its structure. Based on basic syntax patterns: <subject> <verb> <object> <adverbial>, the identifier identifying a semantic relation may be located within any of the <subject>, <verb>, <object>, or <adverbial> components. When refining the step patterns of each semantic relation at the syntactic level, this paper analyzes the syntactic roles in which the identifier identifying the relation may be located. For instance, in steps with a purpose-action relation, the purpose identifier typically appears in the subject, verb, and adverbial positions. Specifically, the purpose identifier can be integrated with other words to serve as subjects (e.g., the noun phrase "the purpose of …" as subject), or integrated with other words, phrases or clauses to serve as adverbials (e.g., the prepositional phrase "in order to <verb phrase>" as adverbial, or the adverbial clause of purpose "in order that <clause>"). Additionally, the purpose identifier can also stand independently as adverbials (e.g., the prepositional phrase "to the end" as adverbial), or as verbs (e.g., the verb phrase "aim to" as verb). Although from a syntactic perspective, the purpose identifier could potentially appear in the object position, such usage is not common or is rarely found in scientific paper writing.

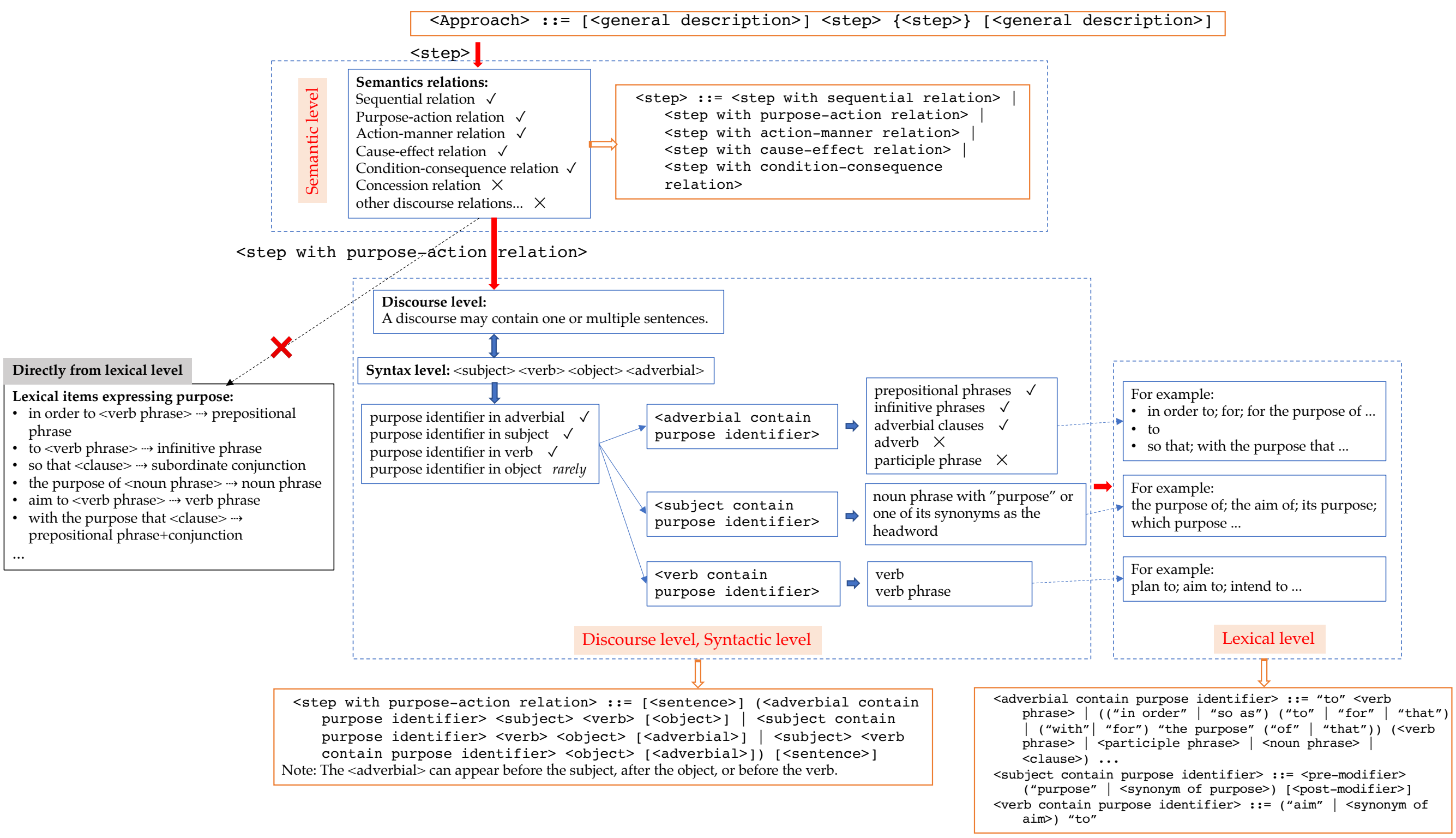

Figure 1. The visual refinement process of approach pattern

After determining which syntactic roles the identifiers expressing a certain semantic relation can appear in, the next step is to identify the lexical items that express that relation based on the part of speech that matches the corresponding syntactic role. Additionally, synonyms that convey the same meaning are also taken into consideration. For example, the purpose identifiers that appear in adverbial position typically consist of prepositional phrases (e.g., in order to <verb phrase>, for the purpose of <participle phrase>, etc.), infinitive phrases (e.g., to <verb phrase>), and adverbial clauses (e.g., so that <clause>, with the purpose that <clause>, and so on).

According to the five discourse relations that can utilized to express steps, the pattern of step can be refined as follows:

```
<step> ::= <step with sequential relation> |
      <step with purpose-action relation> |
      <step with action-manner relation> | <step
      with cause-effect relation> | <step with
      condition-consequence relation>
```

## 2.2 Pattern of Step

### 2.2.1 Pattern of Step with Sequential Relation

Sequential relation of steps refers to the temporal order or sequence in which various steps are executed within a process. In most cases, the execution of a step depends on the successful completion of the preceding step. The sequential relation can be identified through lexical items that denote temporality or sequence, such as "firstly", "then", "the last step", and "after", among others, which are referred to as identifiers.

For example, an approach containing three steps in paper [5] is as follows: "[*The first step*] *involves training a neural network to recognize the type of sentences that should be included in the summary.* [*The second step*], *feature fusion, prunes the neural network and collapses the hidden layer unit activations into discrete values with frequencies…*". The lexical item within "[]" explicitly indicates the sequence nature of the steps, where the words "first" and "second" identify the sequence nature, and the term "step" explicitly identifies the step nature. In addition, the contents of the two listed steps indicate that the second step builds upon the completion of the first, as it relies on the utilization of the previously trained "neural network".

Another approach containing four steps in paper [6] is as follows: "*We* [*first*] *encode each document in the document set and the summary into hidden representations.* [*Then*], *for each document, we select essential sentences by centrality to build a pseudo reference.* [*Next*], *…*". The lexical item within "[]" explicitly indicates the sequence nature of steps.

By analyzing the part-of-speech features of words and the syntax features of phrases that express sequence or temporality, it becomes evident that these words or phrases can be employed in various syntactic roles. They can be integrated with other words to serve as subjects or objects (e.g., the noun phrase "the first step" as subject), or serve as adverbials (e.g., the prepositional phrase "in the first step" as adverbial). Additionally, they can also stand independently as adverbials (e.g., the adverb "firstly" or the adverb phrase "at last" as adverbial). Furthermore, by generalizing patterns from a small set of examples of steps with sequential relations, we discover that the identifiers of such steps can be situated either within the <subject> or the <adverbial> components, i.e.,

```
<step with sequential relation> ::= <step with
      sequential relation identified by subject>
      | <step with sequential relation identi-
      fied by adverbial> | <step with sequential
      relation identified by verb>
```

(1) Step defined by subject containing sequence identifier

The syntax pattern of this type of steps is as follows:

```
<step with sequential relation identified by
      subject> ::= [<adverbial>] <subject con-
      tain sequence identifier> <verb> <object>
      [<adverbial>] {<sentence>}
```

In English grammar, the subject of a sentence is typically represented by a nominal element, such as a noun, pronoun, or noun phrase. Words that express sequence or temporality are typically not nouns. Therefore, they should be used in conjunction with other nouns when serving as part of a subject. By analyzing examples of steps within this type, it becomes evident that sequence words are typically used in conjunction with words explicitly expressing step meaning when they serve as part of a subject. The refined pattern of the subject for this type of steps is as follows:

```
<subject contain sequence identifier> ::=
      (<pre-modifier> <identifier about se-
      quence-adjective> <identifier about step-
      noun> | <identifier about step-noun> <nu-
      meral word>) [<post-modifier>]
```

, where the <identifier about sequence-adjective> corresponds to the sequence words that can be used as adjectives to modify nouns. Examples include "first", "next", "last", and other similar terms. The <identifier about step-noun> corresponds to the nouns that carry the meaning of step, including "step" and its synonyms. The <numeral word> corresponds to the words that convey the concept of number, including terms like "one", "two", "1", "II", and other similar terms.

From a linguistic point of view, verbs can be classified into lexical verbs, linking verbs, auxiliary verbs, and modal verbs. Lexical verbs, also known as main verbs or full verbs, are verbs that have semantic content and carry the primary meaning in a sentence. They express actions, processes, or states. The verb of steps are typically lexical verbs that are used to express actions or operations. Additionally, when the subject of a sentence contains the combination of sequence words and step words, the verb of the step can be linking verbs, e.g., the example "the first step is to …". The refined pattern of this type of steps is as follows:

```
<step with sequential relation identified by
      subject> ::= [<adverbial>] <subject con-
      tain sequence identifier> (<be> ("to" |
      "that") | ("involve" | <synonym of in-
      volve>) | <lexical verb>) (<subordinate
      clause> | <verb phrase> | <participle
      phrase> | <noun phrase>) [<adverbial>]
      {<sentence>}
```

(2) Step defined by adverbial containing sequence identifier

The syntax pattern of this type of steps is as follows:

```
<step with sequential relation identified by
      adverbial> ::= (<adverbial contain se-
      quence identifier> <subject> <verb> [<ob-
      ject>] | <subject> <adverbial contain se-
      quence identifier> <verb> [<object>] |
      <subject> <verb> [<object>] <adverbial
      contain sequence identifier>) {<sentence>}
```

From a linguistic perspective, the adverbial in a sentence can take the form of an adverb (phrase), prepositional phrase, infinitive phrase, participial phrase, or an adverbial clause. By the analysis of examples of steps within this type, it becomes evident that the adverbial containing sequence or temporality identifier can manifest as an adverb (phrase), prepositional phrase, or adverbial clause. Using lexical items to refine the pattern of the adverbial in this type of step, the refined pattern is as follows:

```
<adverbial contain sequence identifier> ::=
      ("in" | "on" | "within" | "for" | "at")
      (("the" | "its" | "this") <identifier
      about sequence-adjective> <identifier
      about step-noun> | <identifier about step-
      noun> <numeral word>) | <identifier about
      sequence-adverb> | <identifier about tem-
      poral-adverb> (<noun phrase> | <participle
      phrase> | <subordinate clause>)
```

, where the <identifier about sequence-adverb> corresponds to the sequence words or phrases that can be used as adverbial to modify sentences. Examples include "firstly", "then", "finally", "at last", and other similar terms. The <identifier about temporal-adverb> corresponds to words or phrases that are typically combined with other phrases or clauses to convey a temporal meaning, encompassing terms such as "after," "once," "when," and other similar expressions.

(3) Step defined by verb containing sequence identifier

The syntax pattern of this type of steps is as follows:

```
<step with sequential relation identified by
      verb> ::= [<adverbial>] <subject> <verb
      contain sequence identifier> <object>
      [<adverbial>] {<sentence>}
<verb contain sequence identifier> ::= "start"|
      "start with"| "end up with"| "initialize"
```

#### *2.2.2 Pattern of Step with Purpose-Action Relation*

If, within a step, one sentence or sentence element describes a purpose and another details the action or operation taken to achieve it, then there exists a purpose-action relation within that step. The purpose-action relation can be identified by analyzing lexical items that explicitly denote a purpose or those that signify an action undertaken to fulfill a specific purpose, such as "in order to", "for that purpose", and "to this end", among others. In addition, the steps with purpose-action relation can be expressed either concisely within a single sentence or elaborated upon across multiple sentences.

For example, one step in an approach outlined in paper [5], which exhibits a clear purpose-action relation, is as follows: "*A neural network is trained* [*to*] *learn the relevant characteristics of sentences that should be included in the summary of the article.*". The lexical item "*to*" within "[]" explicitly denotes the purpose-action relation, with the content following "[*to*]" delineating the purpose and the content preceding "[*to*]" describes the action. This step encapsulates both purpose and action within a single sentence.

Another instance involves a step from a different approach in paper [5] "*We use a conjugate gradient method where the energy function is a combination of error function and a penalty function.* [*The goal of training*] *is to search for the global minima of the energy function.*", where the action and purpose are articulated across two consecutive sentences. In this step, the lexical item within "[]" explicitly identifies the purpose-action relation, wherein the subsequent content delineates the purpose, and the preceding content describes the action.

By analyzing the part-of-speech features of words and the syntax features of phrases that express the purpose, it becomes evident that these words or phrases can be effectively employed in various syntactic roles. They can be integrated with other words to serve as subjects (e.g., the noun phrase "the purpose of …" as subject), or integrated with other words, phrases or clauses to serve as adverbials (e.g., the prepositional phrase "in order to <verb phrase>" as adverbial, or the adverbial clause of purpose "in order that <clause>"). Additionally, they can also stand independently as adverbials (e.g., the prepositional phrase "to the end" as adverbial), or as verbs (e.g., the verb phrase "aim to" as verb). The pattern of <step with purpose-action relation> can be refined as follows:

```
<step with purpose-action relation> ::= <step
      with purpose-action relation identified by
      adverbial> | <step with purpose-action re-
      lation identified by subject> | <step with
      purpose-action relation identified by
      verb>
```

(1) Step defined by adverbial containing purpose identifier

Expressing purpose through adverbials is the most common type of steps with a purpose-action relation. This type of steps encompasses expressing both the purpose and the action within a single sentence, or separating them into two distinct sentences. The syntax pattern of this type of steps is as follows:

```
<step with purpose-action relation identified
```

```
by adverbial> ::= <subject> <verb> [<object>] <adverbial contain purpose identifier> | <adverbial contain purpose identifier> <subject> <verb> [<object>] | <sentence> <adverbial contain purpose identifier> <subject> <verb> [<object>]
```

In a step with a purpose-action relation, if a single sentence expresses both the purpose and the action, the adverbial expresses the entire purpose. However, if the purpose and the action are separated into two distinct sentences, usually the former sentence expresses the purpose, and the main body of the latter sentence expresses the action, with the adverbial in the latter sentence marking the former as its purpose, such as using the phrase *"for that purpose"*.

```
<adverbial contain purpose identifier> ::= <adverbial identifier containing purpose> | <adverbial identifier indicating purpose>
```

, where the <adverbial identifier containing purpose> corresponds to adverbials that contain entire purpose information, and the <adverbial identifier indicating purpose> corresponds to adverbials that do not contain specific purpose information but instead refer to the preceding content as the purpose.

By the analysis of examples of steps within this type, the adverbials that express purpose consist of prepositional phrases, infinitive phrases, and adverbial clauses. Using lexical items to refine the pattern of the adverbial in this type of step, the refined pattern is as follows:

```
<adverbial identifier containing purpose> ::= "to" <verb phrase> | (("in order" | "so as") ("to" | "for") | "for" | ("for" | "with") "the" ("purpose" | <synonym of purpose>) "of") (<verb phrase> | <participle phrase> | <noun phrase>)) | ("so" | "such" | "in order" | "so as" | ("for" | "with") "the" ("purpose" | <synonym of purpose>)) "that" <clause>
<adverbial identifier indicating purpose> ::= "for" ("that" | "the" | "this") "purpose" | ("to" | "for") "this end"
```

(2) Step defined by subject or verb containing purpose identifier

In these two types, the sentence whose subject or verb expresses purpose describes the purpose of a specific step, and the action corresponding to that step usually precedes or follows the sentence. The syntax patterns of these two types of steps are as follows:

```
<step with purpose-action relation identified by subject> ::= <sentence about action> <subject contain purpose identifier> <verb> [<object>] [<adverbial>]
<step with purpose-action relation identified by verb> ::= <sentence about action> <subject> <verb contain sequence identifier> [<object>] [<adverbial>]
```

The <subject contain purpose identifier> is typically a noun phrase with "purpose" or one of its synonyms as the headword, such as "the purpose of <noun phrase>", "its purpose", "which purpose", and so forth. The <verb contain purpose identifier> is typically a verb or verb phrase with the meaning of purpose, such as "aim to", "intent to", "plan to", among others. Using lexical items to refine the pattern of the subject and verb in these two types of steps, the refined patterns are as follows:

```
<subject contain purpose identifier> ::= <pre-modifier> ("purpose" | <synonym of purpose>) [<post-modifier>]
<verb contain purpose identifier> ::= ("aim" | <synonym of aim>) "to"
```

### *2.2.3 Pattern of Step with Action-Manner Relation*

Within a step, if one sentence or sentence element describes the manner or adopted technique in which the action or operation described in another sentence or sentence element has been performed, then an action-manner relation exists within that step. The action-manner relation can be identified by analyzing lexical items that indicate a specific manner or technique, such as "by", "through", "with" and "use", among others.

For example, an approach containing three steps in paper [5] is as follows: "*… The second step, feature fusion, prunes the neural network and collapses the hidden layer unit activations into discrete values with frequencies. This step generalizes the important features that must exist in the summary sentences* [*by*] *fusing the features and finding trends in the summary sentences. The third step, sentence selection,* [*uses*] *the modified* [*neural network*] *to filter the text and to select only the highly ranked sentences…*". For the second step, the lexical item "by" within "[]" explicitly denotes the action-manner relation, with the content following "[by]" delineating the manner and the content preceding "[by]" describes the action. For the third step, the lexical items "uses" and "neural network" within "[]" also denote the action-manner relation, where the main clause of this sentence delineates the manner and the adverbial clause of purpose guided by "to" describes the action.

By analyzing the part-of-speech features of words and the syntax features of phrases that indicate specific manner, it becomes evident that these words or phrases can be employed in multiple syntactic roles. They can be integrated with other words or phrases to serve as adverbials (e.g., the prepositional phrase "by <participle phrase>" as adverbial). Additionally, they can also function as verbs, with their objects typically being an existing technique, method, or model (e.g., the word "use" as verb and "TF-IDF method" as object). Within this case, the action of the step may be described in the sentence preceding or following the manner sentence, or it may be expressed by the adverbials of purpose or the adverbials of manner within the manner sentence, or alternatively, the manner itself can be interpreted as the action. The pattern of <step with action-manner relation> can be refined as follows:

```
<step with action-manner relation> ::= <step with action-manner relation identified by adverbial> | <step with action-manner relation identified by verb>
```

(1) Step defined by adverbial containing manner identifier

Expressing manner through adverbial is the most common type of steps with an action-manner relation. This type of steps typically encapsulates both the manner and the action within a single sentence. The syntax pattern of this type of steps is as follows:

```
<step with action-manner relation identified by
      adverbial> ::= <subject> <verb> [<object>]
      <adverbial contain manner identifier> |
      <adverbial contain manner identifier>
      <subject> <verb> [<object>]
```

By analyzing examples of steps within this type, the adverbials that express manner mainly consist of prepositional phrases and participial phrase. Using lexical items to refine the pattern of the adverbial in this type of step, the refined pattern is as follows:

```
<adverbial contain manner identifier> ::= ("by"
      | "by means of" | "through" | "via" |
      "with" | "without" | "in" | "utilizing" |
      "using") (<participle phrase> | <identifier
      about method-noun>) | "based on" <identi-
      fier about method-noun>) | "in" ["a"|"an"]
      <pre-modifier> ("way"|"fashion"|"manner"|
      <synonym of way>)
```

(2) Step defined by verb containing manner identifier

In this type of step, the main clause of a sentence, where the verb and object jointly articulate the manner of performing an action, conveys the specific manner within the step. The verbs can be certain lexical verbs or verb phrases, such as "use", "make use of", or their synonyms. The objects are typically some existing techniques, methods, or models. The action of the step may be expressed in the preceding or following sentence, or it may be expressed by the purpose adverbials within this sentence, or the manner itself can be interpreted as the action. The syntax pattern of this type of step is as follows:

```
<step with action-manner relation identified by
      verb> ::= [<sentence about action>] (<sub-
      ject> <verb contain manner identifier>
      <object contain manner identifier> | <ob-
      ject contain manner identifier> <be>
      <verb-ed contain manner identifier>) <ad-
      verbial> [<sentence about action>]
```

By analyzing examples of steps within this type, we use lexical items to refine the pattern of the verb and object in this type of step, the refined pattern of the verb and object is as follows:

```
<verb contain manner identifier> ::= "use" |
      "make use of" | "adopt" | "engage" | "uti-
      lize"| "employ" | "choose" | <synonym of
      use>
<object contain manner identifier> ::= <pre-
      modifier> (<identifier about method-noun>
      | "neural network")
```

, where the <identifier about method-noun> corresponds to nouns that carry the meaning of method, encompassing terms like "method", "model", "technique", "framework", "strategy", and so forth, along with their synonyms. The <verb-ed contain manner identifier> corresponds to the past participle form of <verb contain manner identifier>, which, when combined with the auxiliary verb <be>, forms the passive voice.

If the sentence preceding the manner sentence expresses the action of a step, the subject of the manner sentence typically refers to the content mentioned in the preceding sentence, or it may be some certain expressions such as "we", "our model", or other similar terms, to indicate that the approach is proposed within the corresponding scientific paper itself, rather than being proposed by others. Using lexical items to further refine the pattern of the active-voice step within this type, the refined pattern is as follows:

```
<step with action-manner relation identified by
      verb> ::= [<sentence>] ("which" | "it" |
      "we" | ("the" | "this" | "our") ["pro-
      posed"] [<pre-modifier>] (<identifier
      about paper-noun> | <identifier about
      method-noun>) | ("the" | "this" | "these")
      <noun phrase mentioned precedingly>) <verb
      contain manner identifier> <object contain
      manner identifier> [<adverbial contain
      purpose identifier> (<verb phrase> | <par-
      ticiple phrase> | <noun phrase> | <subor-
      dinate clause>)]
```

, where the <identifier about paper-noun> corresponds to nouns that carry the connotation of study, encompassing terms such as "study", "paper", "work", "section" and their synonyms. The <adverbial contain purpose identifier> is introduced in *section 2.2.2*.

### 2.2.4 Pattern of Step with Cause-Effect Relation

One situation involves a step with action-manner relation, where the action of the step can correspond to the result or effect, with the specific manner or technique used in executing that action corresponding to the cause. The identifier that indicates the manner, which serves as the cause, can be some certain prepositions, such as "by", "from", "in", "with", and "without".

Another situation involves a step where the action described in one sentence or sentence element corresponds to the cause, and if that action is executed, it can lead to the result or effect described in another sentence or sentence element. The cause-effect relation can be identified by analyzing lexical items that explicitly denote a result or effect, such as "by doing so", "thereby", and "result in", among others. Steps of this type typically span across two sentences or two clauses.

By analyzing the part-of-speech features of words and the syntax features of phrases that express the effect, it becomes evident that these words or phrases can be effectively employed in various syntactic roles. They can serve as adverbials (e.g., the prepositional phrase "by doing this" serving as an adverbial), or as verbs (e.g., the verb phrase "result in" serving as a verb). In the latter case, the actions of the steps are typically described in the sentence preceding the sentence stating the effect. The pattern of <step with cause-effect relation> can be refined as follows:

```
<step with cause-effect relation> ::= <step
```

```
      with cause-effect relation identified by
      adverbial> | <step with cause-effect rela-
      tion identified by verb>
```

(1) Step defined by adverbial containing effect identifier

Typically, within a step with cause-effect relation, when the adverbial expresses the meaning of an effect or result, its main clause describes the specific content of that effect or result, and often, the sentence preceding the effect sentence details the action of that step. The syntax pattern of this type of steps is as follows:

```
<step with cause-effect relation identified by
      adverbial> ::= <sentence about action>
      <adverbial contain effect identifier>
      <subject> <verb> [<object>] | <subject>
      <verb> [<object>] <adverbial contain ef-
      fect identifier>
```

By analyzing examples of steps within this type, the adverbials that express effect mainly consist of prepositional phrases and adverbs. Using lexical items to refine the pattern of the adverbial in this type of step, the refined pattern is as follows:

```
<adverbial contain effect identifier> ::= "by
      doing" ("so" | "this") | "thereby" | <syn-
      onym of thereby> | "as a result" | "in re-
      sult" | "resulting in" | ("such that"| "so
      that") <clause>
```

(2) Step defined by verb containing effect identifier

In this type, the sentence whose verb expresses the meaning of an effect describes the effect of a specific step, and the action corresponding to that step usually precedes that sentence. The syntax pattern of this type of steps is as follows:

```
<step with cause-effect relation identified by
      verb> ::= <sentence about action> [<adver-
      bial>] <subject> <verb contain effect
      identifier> [<object>] [<adverbial>]
```

The <verb contain effect identifier> is typically a verb or verb phrase with the meaning of effect or result, such as "lead to", "result in", among others. Using lexical items to refine the pattern of the verb, the refined pattern is as follows:

```
<verb contain effect identifier> ::= "result
      in" | "lead to" | <synonym of result in> |
      "have" ("a"|"an") <pre-modifier> ("impact"
      | <synonym of impact>)"on"
```

Typically, because the sentence preceding the effect sentence expresses the action of the same step, the subject of the effect sentence typically refers to the content mentioned in the preceding sentence, such as "it", "this", and "which", among others. By analyzing examples of steps within this type, we use lexical items to further refine the pattern of the step, the refined pattern is as follows:

```
<step with cause-effect relation identified by
      verb> ::= <sentence> ("it" | "this" |
      "these" | "which" | "the" <noun phrase
      mentioned precedingly>) <verb contain ef-
      fect identifier> [<object>] [<adverbial>]
```

### *2.2.5 Pattern of Step with Condition-Consequence Relation*

If, within a step, a sentence or sentence element describes an action that can be successfully executed once a specified condition, described in another sentence or sentence element, is met, then a condition-consequence relation exists within that step. One situation involves a step containing an action-manner relation, where the specific manner or technique employed can serve as a condition for successfully executing the action.

Another common situation involves a step where a specified condition, indicated by a condition identifier like "if", must be met for an action to be successfully executed. The condition-consequence relation can be identified by analyzing lexical items that explicitly denote a condition, such as "if", "as long as", and "once", among others.

For example, one step in an approach outlined in paper [5], which contains a clear condition-consequence relation, is as follows: "*Second, we utilize the sentence centrality computed from sentence-level representations of the source document to produce the importance weights of the pseudo reference sentences and tokens.* [*Based on*] *the weights, we compute a weighted relevance score that is more precise by considering the relative importance.*". The lexical item "*based on*" within "[]" explicitly denotes the condition-consequence relation, with the adverbial guided by "[*based on*]" delineating the condition and the main clause of the second sentence describes the action.

Another instance involves a step from a different approach in paper [5] "*We use a three-layered feedforward neural network, which has been proven to be a universal function approximator. It can discover the patterns and approximate the inherent function of any data to an accuracy of 100%,* [*as long as*] *there are no contradictions in the data set.*". In this step, the lexical item "as long as" within "[]" explicitly identifies the condition-consequence relation, with the content following "[*as long as*]" delineating the condition and the content preceding "[*as long as*]" describes the action.

By analyzing the part-of-speech features of words and the syntax features of phrases that indicates the condition, it becomes evident that these words or phrases are typically used alongside other words, phrases or clauses to function as adverbials. For example, the adverbial clause of condition "if <clause>" functions as an adverbial, or the prepositional phrase "with <participle phrase>" also serves as an adverbial.

(1) Step defined by adverbial containing condition identifier

Expressing condition through adverbials is the most common type of steps with a condition-consequence relation. This type of steps typically encapsulates both the condition and the action within a single sentence, with the adverbial clause of condition expressing the condition and the main clause expressing the action. The syntax pattern of this type of steps is as follows:

```
<step with condition-consequence relation> ::=
      <adverbial contain condition identifier>
      <subject> <verb> [<object>]
```

By analyzing examples of steps within this type, the adverbials that express condition mainly consist of prepositional phrases and adverbial clauses. Using lexical items to refine the pattern of the adverbial in this type of step, the refined pattern is as follows:

```
<adverbial contain condition identifier> ::=
      (“as long as” | “if” | “if and when” |
      “once” | “until” | “when” | “whenever” |
      “in case of”) <subordinate clause> |
      (“based on” | “upon”) (<participle phrase>
      | <noun phrase>)
```

Overall, the step patterns of these five relations are confirmed through the analysis of steps identified in scientific papers’ approaches.

## 2.3 Verification of Approach Pattern

### *2.3.1 Experimental Dataset*

Due to the lack of publicly available datasets for extracting complete appraoches from scientific papers, we collected scientific papers covering various topics and manually annotated approaches. Specifically, the constructed dataset consists of two parts: one part involves extracting approaches from the whole paper, while the other focuses on extracting approaches from specific sections (excluding related work and experimental sections). The entire annotation process includes one round of initial annotation and one round of correction and improvement.

Table 1 shows the details of the dataset for appraoch extraction from whole papers. We selected a total of 20 papers from four different topics (5 papers per topic), and annotated approaches throughout the full texts of these papers. The topics include text summarization, text clustering, machine translation, and frequent itemset mining.

Table 1. Introduction of dataset for extracting approaches from full paper

| Topic | Num of approaches | Num of steps |
|---|---|---|
| Text Summarization | 69 | 210 |
| Text Clustering | 61 | 159 |
| Machine Translation | 67 | 188 |
| Frequent Itemset | 77 | 214 |
| Total | 274 | 771 |

During the annotation process, we observed that approaches extracted from related work sections were mostly from previous studies, rather than the research approaches discussed in the current paper. In addition, the experimental section not only includes the presentation and analysis of experimental results, but also covers the design approaches and procedures of the experiments. In general, the design process of experiments can also be considered as an approach, describing how the experiment was conducted. Typically, researchers are more focused on approaches that solve specific problems or achieve particular goals, rather than on experimental design appraoches. Experimental design approaches in papers usually provide detailed descriptions of the experimental setup, steps, and procedures, and are primarily used to validate and evaluate the effectiveness of the research approaches that address the targeted problems.

In order to focus on the approaches proposed in scientific papers for solving specific scientific problems or achieving particular goals, we extracted approaches from sections of the papers other than the related work and experimental sections. Table 2 presents the details of the dataset for approach extraction from specific sections. Specifically, we selected 10 papers from each of the following four topics (a total of 40 papers): text summarization, text clustering, machine translation, and frequent itemset mining. In addition, we selected 10 papers from other topics. Approach annotations were conducted on specific sections of these papers.

Table 2. Introduction of dataset for extracting approaches from particular sections

| Topic | Num of approaches | Num of steps |
|---|---|---|
| Text Summarization | 88 | 278 |
| Text Clustering | 105 | 255 |
| Machine Translation | 115 | 322 |
| Frequent Itemset | 112 | 292 |
| Others | 62 | 219 |
| Total | 482 | 1366 |

### *2.3.2 Experimental Results*

This section evaluates the precision and completeness of the approach patterns by testing precision, recall, and F1-Score for the approach extraction task on datasets.

Tables 3 and Table 4 show the results of approach extraction from full papers and from specific sections respectively, covering the overall precision, recall, and F1-Score metrics for approach and step extraction across all topics. An approach is considered successfully matched by a pattern if at least one of its steps can be matched by the pattern.

Table 3. Overall results for approach extraction from full paper

| | Precision | Recall | F1-Score |
|---|---|---|---|
| Step | 94.38 | 93.64 | 94.01 |
| Approach | 95.39 | 98.18 | 96.76 |

Table 4. Overall results for approach extraction from particular sections

| | Precision | Recall | F1-Score |
|---|---|---|---|
| Step | 95.88 | 93.78 | 94.82 |
| Approach | 97.55 | 98.96 | 98.25 |

We can observe that the results for approach matching are better than those for step matching. This is because most of the steps that failed to be matched were actually part of approaches containing multiple steps. According to our evaluation criteria, if at least one step within an approach is successfully matched, the entire approach is considered to be matched. Therefore, even if some steps are not correctly identified, the corresponding approach may still be marked as successfully matched, which improves the overall approach matching performance to some extent.

#### 2.3.3 Discussion

The multi-level approach patterns constructed in this paper have irreplaceable advantages, but they also have some limitations.

One major advantage of the multi-level approach patterns is their extensibility. Although the patterns constructed in this paper have already achieved high completeness and accuracy, there are still some approaches or steps that cannot be matched, as well as cases where non-step sentences are mistakenly identified as steps. For the approaches or steps that were not matched, the point of failure can be analyzed at different levels—such as the semantic level, discourse level, syntactic level, or lexical level—and the patterns can then be improved and expanded accordingly. For example, the reason that step "*The English output sentence is generated left to right [in form of] partial translations (or hypotheses).*" was not matched by any pattern was due to a lack of coverage at the lexical level. This can be addressed by adding specific lexical items into the corresponding lexical-level patterns, thereby expanding the approach patterns so that similar expressions encountered in the future can be successfully matched. However, it is not necessary to extend the patterns for every unmatched step instances. This is because simply expanding the patterns, while increasing coverage, may also lead to mistakenly identifying many non-step sentences that contain the same relation as steps, thereby reducing the precision of the patterns.

For sentences that are incorrectly matched, the precision of the patterns can also be improved by adding some constraints. For example, by introducing heuristic constraints such as requiring that the predicate verb in a sentence expressing an action cannot be in the perfect tense, non-step sentences like "*Many text mining methods have been developed [to] retrieve aggregated information that is useful for users.*" and "*Agglomerative clustering methods have been widely used by many research communities [to] cluster their data into hierarchical structures.*" can be prevented from being misidentified as steps. However, it should be noted that although adding such heuristic rules can improve the precision of pattern matching, they may also lead to the exclusion of genuine steps that happen to conform to these constraints, thereby reducing the overall recall of the patterns.

Therefore, when expanding the multi-level approach patterns through examples, it is important to strike a balance between precision and recall. This means that while adding new rules or adjusting existing ones to improve matching precision, one must also consider that these changes may cause some valid steps to be overlooked, thereby reducing the overall coverage of the patterns. Conversely, when expanding the patterns to increase coverage, it is equally important to recognize that such modifications may lead to some non-step sentences being incorrectly identified as steps, thus lowering the overall precision of the patterns.

## 3 Constructing Multiple Dimensions for Approach

### 3.1 Dimensions of Approach

Action is the most fundamental element of a step, clearly indicating what is done in that step. In addition, some steps may include purpose element, which explains the intention behind performing the step through a Purpose-Action relation. Steps may also contain manner element, describing the specific manner or technique used to carry out the action through an Action-Manner relation. Furthermore, steps can include condition and effect elements, which explain under what conditions the step is performed via a Condition-Consequence relation, and what outcomes result from performing the action via a Cause-Effect relation, respectively.

Table 5 presents the distribution of the five elements of approaches (action, purpose, manner, condition, and effect) in the self-built dataset. As shown in the table, all approaches contain action element, since action is the most fundamental element of a step. Over 70% of the approaches include both the purpose and manner elements, indicating that most approaches in scientific papers not only focus on "what is done" but also clearly express the purpose and the specific techniques or manners used. However, the coverage of the condition and effect elements is relatively low, reaching only 37% and 27%, respectively. This is because, in actual writing, the description of preconditions or expected outcomes of approaches or steps is often not as explicit as that of actions, purposes, or manners. In many cases, it is implicitly assumed that the steps of an approach, when successfully executed, will achieve the goal stated in the purpose element. Moreover, in approaches containing multiple steps, the successful completion of one step often serves as the condition for the next step. To avoid redundancy, such dependency relationships are typically not explicitly expressed in scientific papers.

Table 5. The distribution of five types of elements in approaches

| Element | action | purpose | manner | condition | effect |
|---|---|---|---|---|---|
| Approach | 100 | 70.8 | 74.2 | 37.0 | 27.0 |

Based on the above discussion, this paper generalizes five dimensions of approaches, with each dimension representing a different classification criterion.

- **Action.** Every step contains action element, which expresses what the step specifically does. Action can serve as one dimension of approaches, enabling the approaches with similar actions to be grouped into the same class. The Action dimension is capable of managing all approaches.
- **Purpose.** Purpose is an important element of an approach or step, as it explains the specific problem the approach or step aims to solve or the concrete goal it intends to achieve. We use purpose as a dimension to categorize the approaches, facilitating the grouping of approaches with analogous objectives into the same class. The purpose dimension is able to manage more than 70% of the approaches.
- **Manner.** Another crucial element of an approach or

step is manner, which explains how the action is performed and what specific techniques are used to achieve it. We also employ manner as a dimension, enabling approaches that employ similar techniques or manners to be grouped into the same class. The manner dimension is able to manage more than 74% of the approaches.

- **Condition.** The condition of an approach or step refers to the necessary conditions or requirements that must be satisfied before implementing the approach or step. It serves as an important basis for initiating the approach or step, ensuring a smooth and effective implementation process. The condition of a step within an approach can be explicitly described in paper or implicitly reflected by the preceding step. Condition can also serve as a classification dimension, facilitating the grouping of approaches with analogous condition into the same class. By treating the previous step in an approach as the execution condition for the subsequent step, the condition dimension is able to manage over 70% of the approaches (as shown in Table 6).
- **Effect.** The effect of an approach or step refers to the outcome or consequence that results from implementing that approach or step. The effect may be explicitly presented in paper or implicitly reflected in the resolution of the problem or attainment of the stated objective as outlined in the “purpose”. Effect also serves as a classification dimension, enabling approaches that yield similar effects or outcomes to be grouped into the same class. By treating the purpose element as the actual result for approaches or steps that do not explicitly state a result, the effect dimension is able to manage over 76% of the approaches (as shown in Table 6).

Different dimensions provide diverse perspectives for categorizing and organizing approaches in scientific papers. The approaches that are classified into the same class in one dimension may not fall into the same class in other dimensions, and in fact, they may exhibit significant differences.

Table 6. The coverage of different dimensions to the approaches

| Dimension | Action | Purpose | Manner | Condition | Effect |
|---|---|---|---|---|---|
| Approach | 100 | 70.8 | 74.2 | 70.8 | 76.5 |

## 3.2 Construction of Multiple Dimensions

By utilizing the approach patterns, our model is capable of extracting approaches from scientific papers and identifying the approach components within each dimension. Specifically, each approach comprises multiple steps, with each step further divided into five components: action, purpose, manner, condition, and effect. Each of these components is embodied as a linguistic unit, ranging from a single word or phrase to a clause, a sentence, or even a sequence of sentences. The projection of an approach within a single dimension is a collection of step components, where each step component corresponds to the representation of the step within that dimension. Consequently, each approach is represented by five distinct collections of step components, with each collection corresponding to one dimension and referred to as an approach component within that dimension.

To effectively manage approaches from different dimensions, our model needs to construct each dimension for the approach components within that dimension.

An efficacious way involves constructing class trees for each dimension. Hierarchical classification of approaches can form classes at multiple levels, facilitating organization and management of approaches across different granularities. In class trees, classes closer to the root node are expected to contain a smaller amount of more generalized common semantics shared across more approaches, whereas classes closer to the leaf nodes are expected to encompass a greater quantity of more specific common semantics shared by fewer approaches. Clustering is a way to construct class trees where each node represents a class of approaches. The core of clustering lies in similarity calculation.

Our model introduces innovative algorithms for calculating similarity between approach components and for clustering approach components, respectively.

### 3.2.1 *Similarity Calculation between Approach Components within one Dimension*

Calculating the similarity between two approach componentes refers to calculating the similarity of their respective collections of step components.

Different approaches may contain one or more similar steps. Before calculating the similarity between approach components, it is necessary to first compute the similarity between individual step components from different approach components, so that similar step components across different approach components can be extracted.

**(1) Similarity calculation between step components.**

A step component may consist of a sentence or a part of a sentence (such as a phrase or a word), or even multiple sentences. It is reasonable to account for similarities at the syntactic level when comparing different step components.

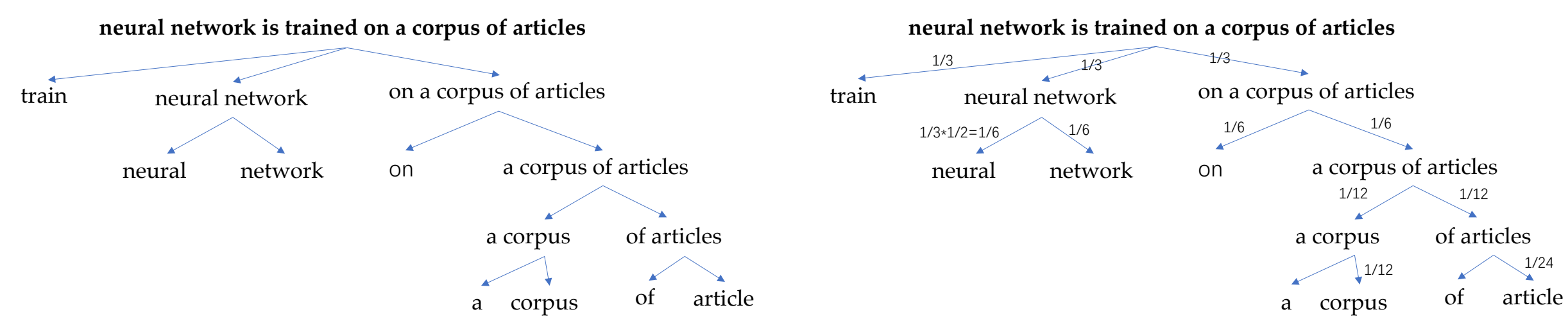

Figure 2. Left: The tree structure for a step component: “*neural network is trained on a corpus of articles*”; Right: The similarity-weights assigned to the nodes in the tree structure.

This paper proposes a novel tree-structure-based similarity measure for calculating the similarity between step components. Firstly, each step component is represented as a tree structure, with the root node representing the step component itself, and the leaf nodes consist of individual word within the step component. Then, calculating the similarity between two step components is transformed into computing the similarity between their respective tree structures.

Our model constructs a tree structure for each step component based on its syntactic structure and syntactic-related information. The detailed procedure is as follows:

**Step 1:** Use the step component itself as the root node.

**Step 2:** Extract the subject element, predicate element, object element, and adverbial element of the step component, and allocate them as the second-level sub-nodes on the tree structure.

The subject element, predicate element, object element, and adverbial element here are not exactly the same as the standard syntactic structure in a sentence. Specifically, the main verb in the predicate of a step component serves as the predicate element, i.e., if the predicate of the step component is in the passive voice, only the main verb is considered part of the predicate element, while the auxiliary verb is excluded. The initiator of the predicate element serves as the subject element, the recipient of the predicate element serves as the object element, and additional information related to the predicate, such as the time or place where the predicate occurs, serves as the adverbial element. That is, if a step component is expressed in the passive voice, it is essential to first identify the true initiator (subject element) and recipient (object element) of the main verb.

**Step 3:** Complete the subtrees rooted at the subject element, predicate element, object element, and adverbial element, respectively, according to their corresponding syntactic parsing tree structures. Specifically, each node in these subtrees is represented by a string formed from the concatenation of all its leaf nodes in the syntactic parsing tree. In addition, within these subtrees, if a node has only one subnode, the subnode can be directly removed, as it is redundant with its parent node. The subnodes of the removed subnode will then become the subnodes of its parent node.

The left side of Figure 2 presents an example of the tree structure for a step component: "*neural network is trained on a corpus of articles*".

After converting each step component into a tree structure, a tree-structure-based similarity measure is designed to compute the similarity between different tree structures, such that (1) if two step components are identical or synonymous, the similarity between their corresponding tree structures is 1; and (2) the closer identical or synonymous nodes are to the leaf nodes on tree structures, the less common information the two step components share, resulting in a lower similarity score.

A **similarity-weight** is assigned to each node in a tree, representing the contribution of that node's similarity score to the overall similarity score of the entire tree. Specifically, if node $a$ in tree $A$ is identical or synonymous with a node in tree $B$, the similarity-weight of node $a$ in tree $A$ reflects its contribution to the overall similarity score from tree $A$ to tree $B$. The similarity-weight for each node in a tree is assigned as follows: (1) the similarity-weight of the root node is set to 1, indicating that if the root node of this tree is identical or synonymous with a node in another tree, the overall similarity score from this tree to the other tree is 1; (2) if a node with $m$ child nodes has a similarity-weight of $p$, then the similarity-weight of each child node is assigned as $p/m$.

The right side of Figure 2 illustrates the similarity-weights assigned to the nodes in the tree structure presented on the left.

To calculate the similarity between two trees, our model traverses these two trees from top to bottom to identify identical or synonymous nodes. If two nodes in the separate trees are identical or synonymous, the subtrees rooted at these nodes are also identical or synonymous. Consequently, once identical or synonymous nodes are identified in two separate trees, their subnodes are excluded from further consideration.

Based on the similarity-weights of the identical or synonymous nodes in two trees $A$ and $B$, the similarity between tree $A$ and tree $B$ is calculated as follows:

$$Tree_similarity(A,B) = \frac{Tsim_{A\to B} + Tsim_{B\to A}}{2}$$
$$= \frac{\sum_{\forall a\in A, \exists b\in B \Rightarrow a\approx b} sw_{a\in A} + \sum_{\forall b\in B, \exists a\in A \Rightarrow b\approx a} sw_{b\in B}}{2} \quad (1)$$

where $sw_{a\in A}$ is the similarity-weight of node $a$ in tree $A$. $Tsim_{A\to B}$ calculates the sum of the similarity-weights of nodes in tree $A$ that are identical or synonymous with at least one node in tree $B$. Likewise, $Tsim_{B\to A}$ calculates the sum of the similarity-weights of nodes in tree $B$ that are identical or synonymous with at least one node in tree $A$. The overall similarity score between tree $A$ and tree $B$ is computed as the average of $Tsim_{A\to B}$ and $Tsim_{B\to A}$.

Overall, if two step components are identical or synonymous, then the root nodes of their respective trees are also identical or synonymous, and their similarity calculated based on formula (1) is given by (1+1)/2=1. If the identical or synonymous nodes are positioned lower in two trees, the similarity between the two trees decreases, as the common information lies in less significant syntactic constituent of the step components.

Figure 3 presents two examples of similarity calculations conducted between different tree structures. In the upper example of Figure 3, the identical nodes are "*train*" and "*neural network*", which are located at the second level of the trees, resulting in relatively higher similarity-weights. In the lower example, the identical node is "*query*", positioned at the last level of the trees, leading to lower similarity-weights compared to the upper example. Furthermore, the tree similarity of the upper example is greater than that of the lower example.

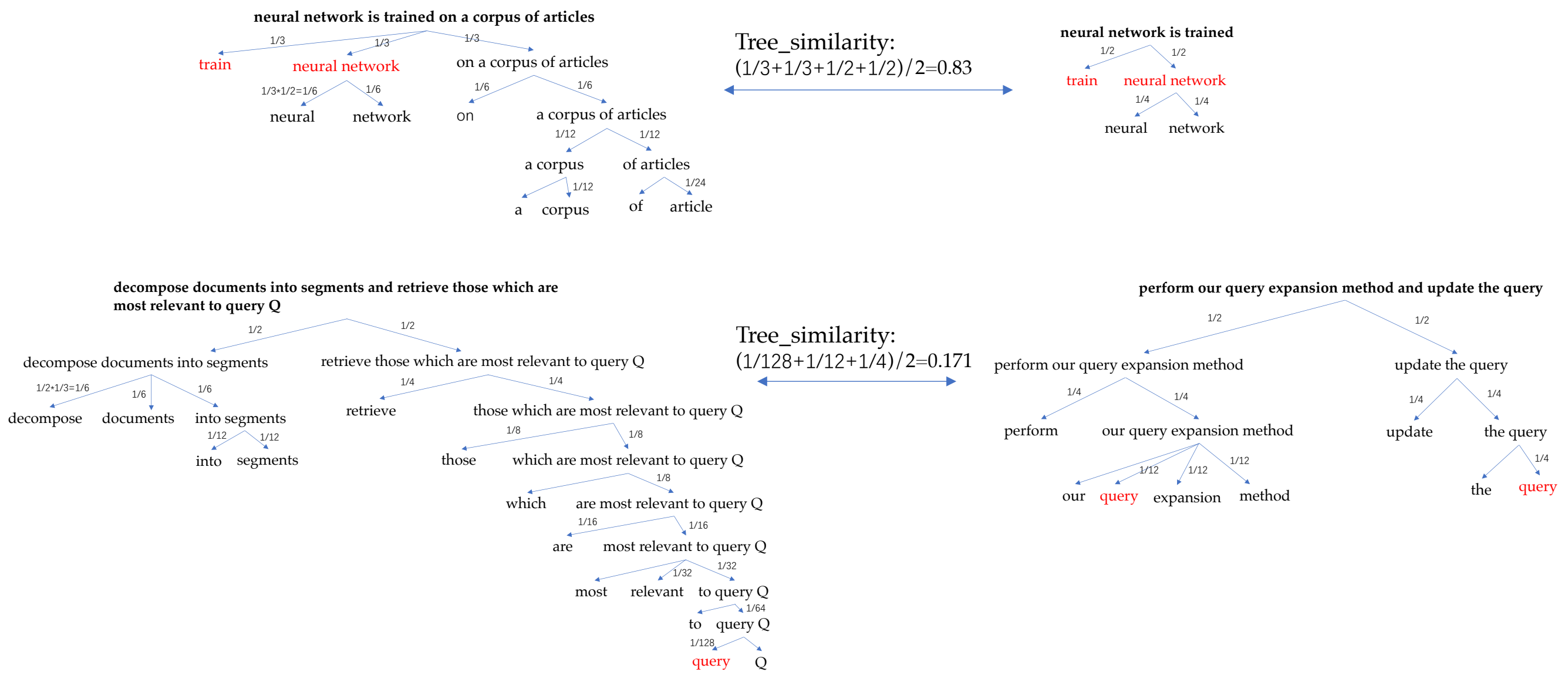

Figure 3. Similarity calculation between different tree structures. The identical nodes in each pair of trees are highlighted in red for clarity.

**(2) Similarity calculation between approach components**

Calculating the similarity between approach components involves computing the similarity between their respective collections of step components. The similarity between collections of step components should capture as much common elements between step components as possible while avoiding redundancy, thereby facilitating the extraction of identical (or synonymous) step components or common elements between step components in distinct approach components. This paper presents an innovative algorithm for calculating the similarity between different approach components.

Formally, for two approach components $AP_1$ and $AP_2$, where $AP_1$ contains three step components: $AP_1 = \{A_1, A_2, A_3\}$, and $AP_2$ contains four step components $AP_2 = \{B_1, B_2, B_3, B_4\}$. Firstly, for each step component $A_i \in AP_1$, compute its tree similarity with each step component $B_j \in AP_2$ using the proposed tree-structure-based similarity measure, and then identify the most similar step component in $AP_2$, denoted as $B^*$. Furthermore, extract the common element shared by $A_i$ and $B^*$ based on their tree structures, which serves as one of the class labels for the class formed by $AP_1$ and $AP_2$. For instance, the common element extracted from the upper example in Figure 3 is "*train neural network*", and the common element extracted from the lower example in Figure 3 is "*query*". Next, calculate the sum of the tree similarities between all step components in $AP_1$ and their most similar step components in $AP_2$ to obtain the overall similarity from $AP_1$ to $AP_2$ (denoted as $sim_{AP_1 \rightarrow AP_2}$). Then, use the same way to calculate the overall similarity from $AP_2$ to $AP_1$ (denoted as $sim_{AP_2 \rightarrow AP_1}$). Finally, the similarity between $AP_1$ and $AP_2$ is calculated as the average of the similarity from $AP_1$ to $AP_2$ and the similarity from $AP_2$ to $AP_1$. The class labels for the class formed by $AP_1$ and $AP_2$ is the union of the common elements shared between each step component in $AP_1$ and its most similar step component in $AP_2$, along with the common elements shared between each step component in $AP_2$ and its most similar step component in $AP_1$.

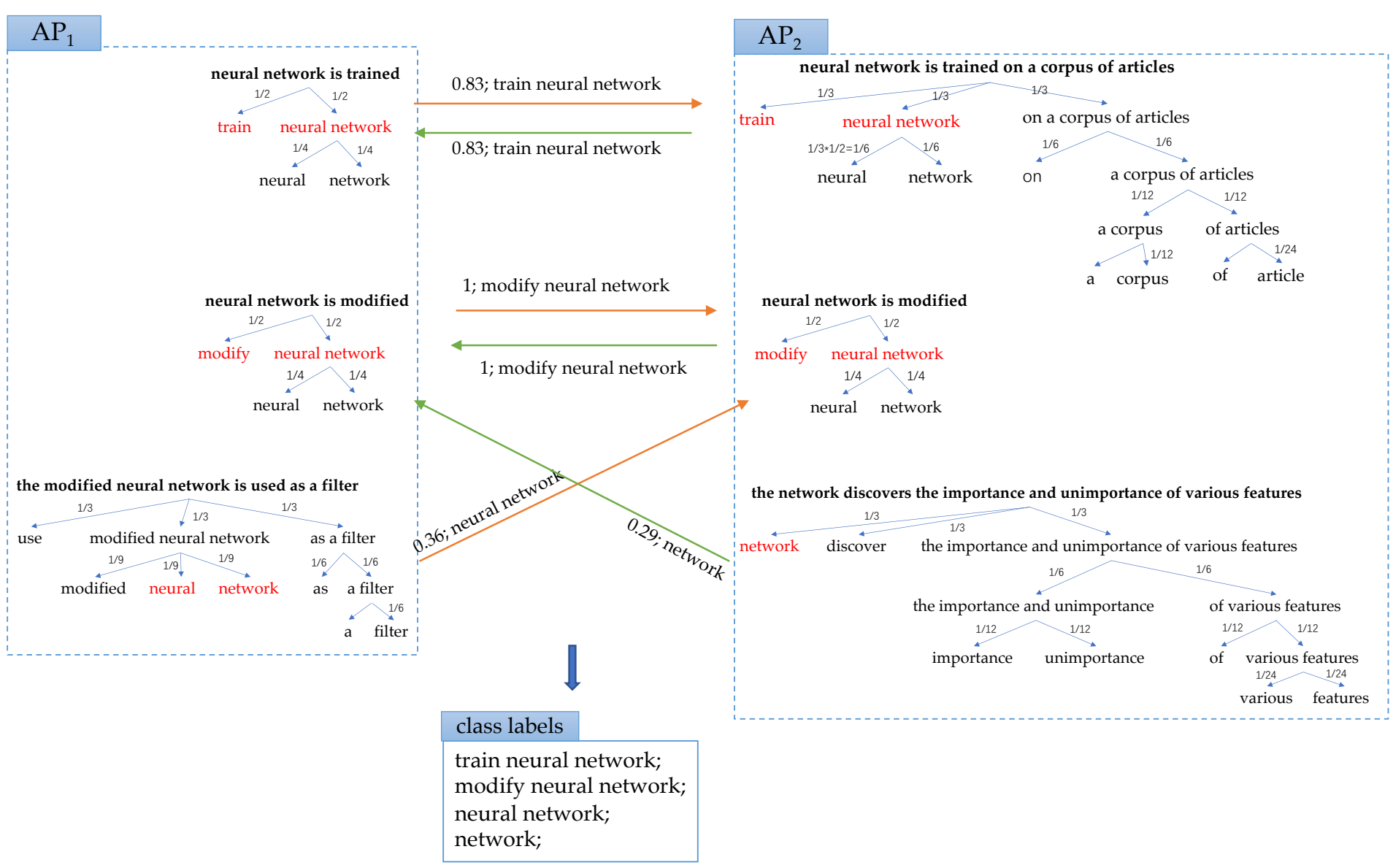

Figure 4. An example of similarity calculation conducted between approach components.

The following is the function for calculating the similarity between two approach components $AP_1$ and $AP_2$:

$$Similarity(AP_1, AP_2) = \frac{sim_{AP_1 \to AP_2} + sim_{AP_2 \to AP_1}}{2}$$

$$sim_{AP_1 \to AP_2} = \sum_{A_i \in AP_1} \max_{B_j \in AP_2} Tree_similarity(A_i, B_j)$$

$$sim_{AP_2 \to AP_1} = \sum_{B_j \in AP_2} \max_{A_i \in AP_1} Tree_similarity(A_i, B_j) \quad (2)$$

Figure 4 presents an example of similarity calculations conducted between approach components. The approach component on the left side ($AP_1$) and the approach component on the right side ($AP_2$) each contain three step components. Each step component in an approach component is linked to its most similar step component in the other approach component based on tree similarity, with the tree similarity score and the corresponding class label presented. For instance, the step component in $AP_2$ that is most similar to the step component "*neural network is trained*" in $AP_1$ is the step component "*neural network is trained on a corpus of articles*", with a tree similarity score of 0.83 and a common element of "*train neural network*". Additionally, the class labels of the class formed by $AP_1$ and $AP_2$ are displayed below Figure 4, and the overall similarity between $AP_1$ and $AP_2$ is computed as (0.83+1+0.36+0.83+1+0.29)/2=2.155.

### 3.2.2 Construction of Classification Dimension based on Clustering

To construct class trees for each dimension, our model clusters approach components within each dimension, such that similar approach components can be grouped into the same class, with the common elements shared among the step components in respective approach components serving as the class labels.

Based on the similarity measure for approach components, an algorithm can be designed for clustering approach components into class trees such that (1) one approach component could be clustered into multiple classes because one approach component may share different common elements between its step components and those of different approach components; (2) a father class contains the approach components of its subclasses; and, (3) a father class has fewer or more concise class labels than its subclasses because a subclass contains less approach components.

The proposed algorithm for constructing class trees consists of the following steps, which inputs a set of approach components $APs = \{AP_1, AP_2, \ldots, AP_n\}$ and then outputs a set of class trees:

**Step 1**: Each approach component $AP_i$ serves as an individual class and corresponds to a leaf node in the class trees, with the step components within $AP_i$ serving as its class labels.

**Step 2**: For each approach component $AP_i \in APs$, the most similar approach component $AP_j$ ($AP_j \in APs$ and $i \neq j$) is identified to form a class containing both $AP_i$ and $AP_j$. All formed classes are reserved so that each approach component meets its most similar approach component.

The class labels of the class formed by $AP_i$ and $AP_j$ is the union of the common elements shared between each step component in $AP_i$ and its most similar step component in $AP_j$, along with the common elements shared between each step component in $AP_j$ and its most similar step component in $AP_i$. The weight associated with each class label (i.e., each entry in the class labels) is derived from their similarity-weights in both $AP_i$ and $AP_j$, and the total weights of all class labels reflect the similarity score of the class. In the example illustrated in Figure 4, the weights for the class labels are calculated as follows: the weight for "*train neural network*" is computed as (0.83+0.83)/2=0.83; the weight for "*modify neural network*" is (1+1)/2=1; the weight for "*neural network*" is 0.36/2=0.18; and the weight for "*network*" is 0.29/2=0.145.

**Step 3**: For each class formed in step 2 (denoted as $class_p$), identify the most similar class (denoted as $class_q$, and $p \neq q$) to form a new class that contains all the approach components contained in both $class_p$ and $class_q$ based on the similarity score calculated between the two classes.

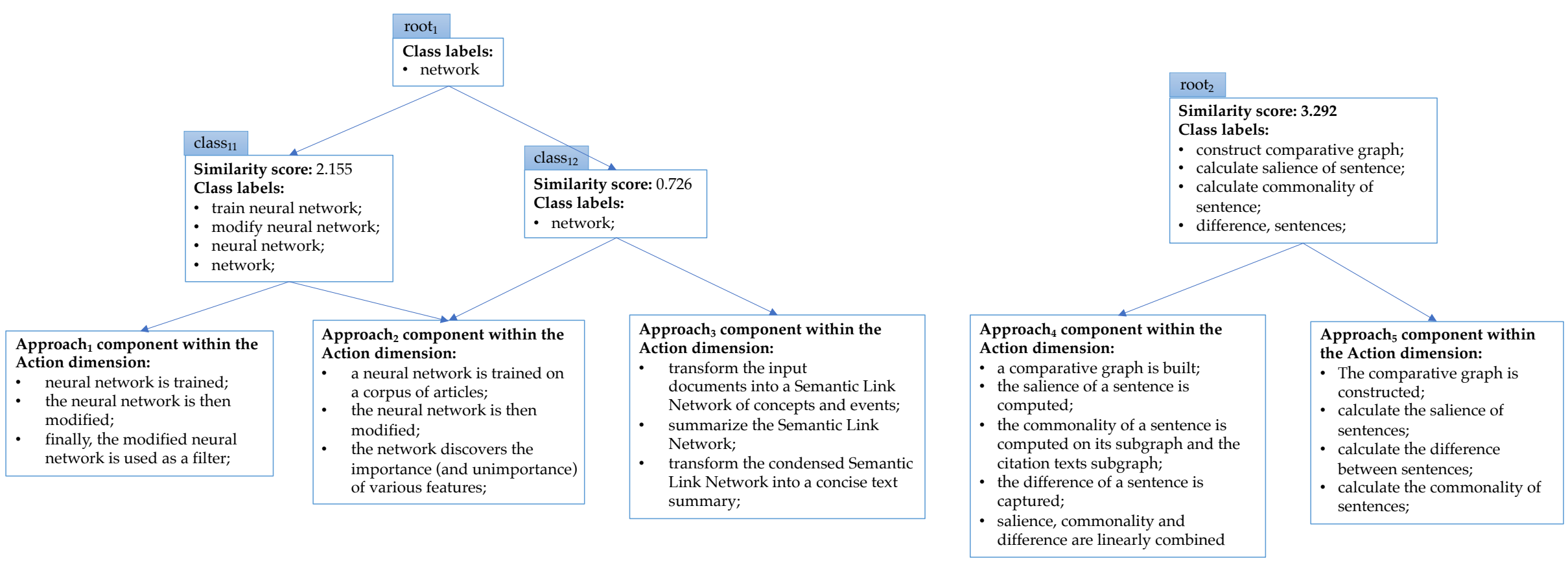

Figure 5. An example of class trees on the Action dimension

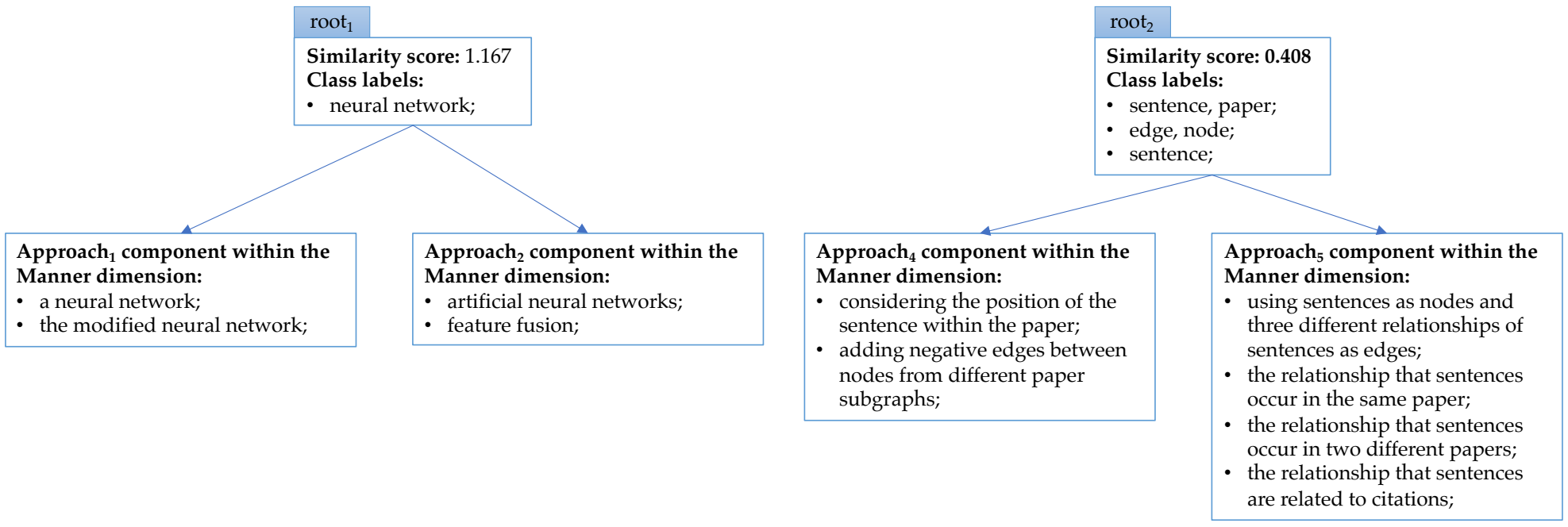

Figure 6. An example of class trees on the Manner dimension

Our model treats each class as a new approach component when calculating the similarity between classes. Specifically, each class label of the class is considered a step component, with its weight serving as its similarity-weight when calculating tree similarity with other class label. The similarity between two classes is computed using Formula (2).

The class labels for the new class are the union of the common elements shared between each entry in the class labels of $class_p$ and its most similar entry in the class labels of $class_q$, along with the common elements shared between each entry in the class labels of $class_q$ and its most similar entry in the class labels of $class_p$. The weight associated with each new class label (i.e., each entry in the class labels of the new class) is derived from their similarity-weights in both $class_p$ and $class_q$. The total weights of all new class labels reflect the similarity score of the new class.

**Step 4**: Repeat Step 3 until no further classes can be merged. Classes without shared common elements cannot be merged.

**Step 5**: Output the class trees with the class labels for each class, which constitute a classification on the input set of approach components *APs*.

Figure 5 shows an example of the class trees built on the action dimension, with two trees constructed in total. In the class trees, class labels become increasingly concise and general as one moves from leaf nodes to root nodes. Additionally, during each step of the clustering process, the algorithm identifies the most similar class for each class and merges them, unless the class shares no common step elements with any other class. For example, although Approach$_3$ does not belong to the same class as Approach$_1$ and Approach$_2$, it was merged with Approach$_2$ because they both contain the term "network" and are thus more similar. However, as shown in the figure, the similarity within class$_{11}$, composed of Approach$_1$ and Approach$_2$, is significantly higher than that within class$_{12}$, composed of Approach$_2$ and Approach$_3$. Moreover, the algorithm did not merge root$_2$ with either class$_{11}$ or class$_{12}$, since root$_2$ shares no common elements with the labels of class$_{11}$ or class$_{12}$.

The procedure for constructing different classification dimensions remains the same, with the only variation being the input. For instance, the input for constructing the action dimension is the approach components within the action dimension, while the input for constructing the purpose dimension is the approach components within the purpose dimension. Figure 6 and Figure 7 show the class trees built for the same set of approaches on the manner dimension and the purpose dimension, respectively. Specifically, Figure 6 presents the results on the manner dimension. Since Approach$_3$ does not have component in the manner dimension, it was not included in the constructed class tree. Two class trees were built on the manner dimension: Approach$_1$ and Approach$_2$ were grouped into one tree, while Approach$_4$ and Approach$_5$ were grouped into another. Figure 7 shows the results on the purpose dimension. Although both the action dimension and the manner dimension classify $Approach_1$ and $Approach_2$ into the same class, and $Approach_4$ and $Approach_5$ into another class, these approaches ($Approach_1$, $Approach_2$, $Approach_4$, and $Approach_5$) are ultimately grouped into the same class tree under the purpose dimension. Meanwhile, $Approach_3$ is not grouped with any other approach. Therefore, approaches that are classified into the same class under some dimensions may not belong to the same class under other dimensions and may even exhibit significant differences.

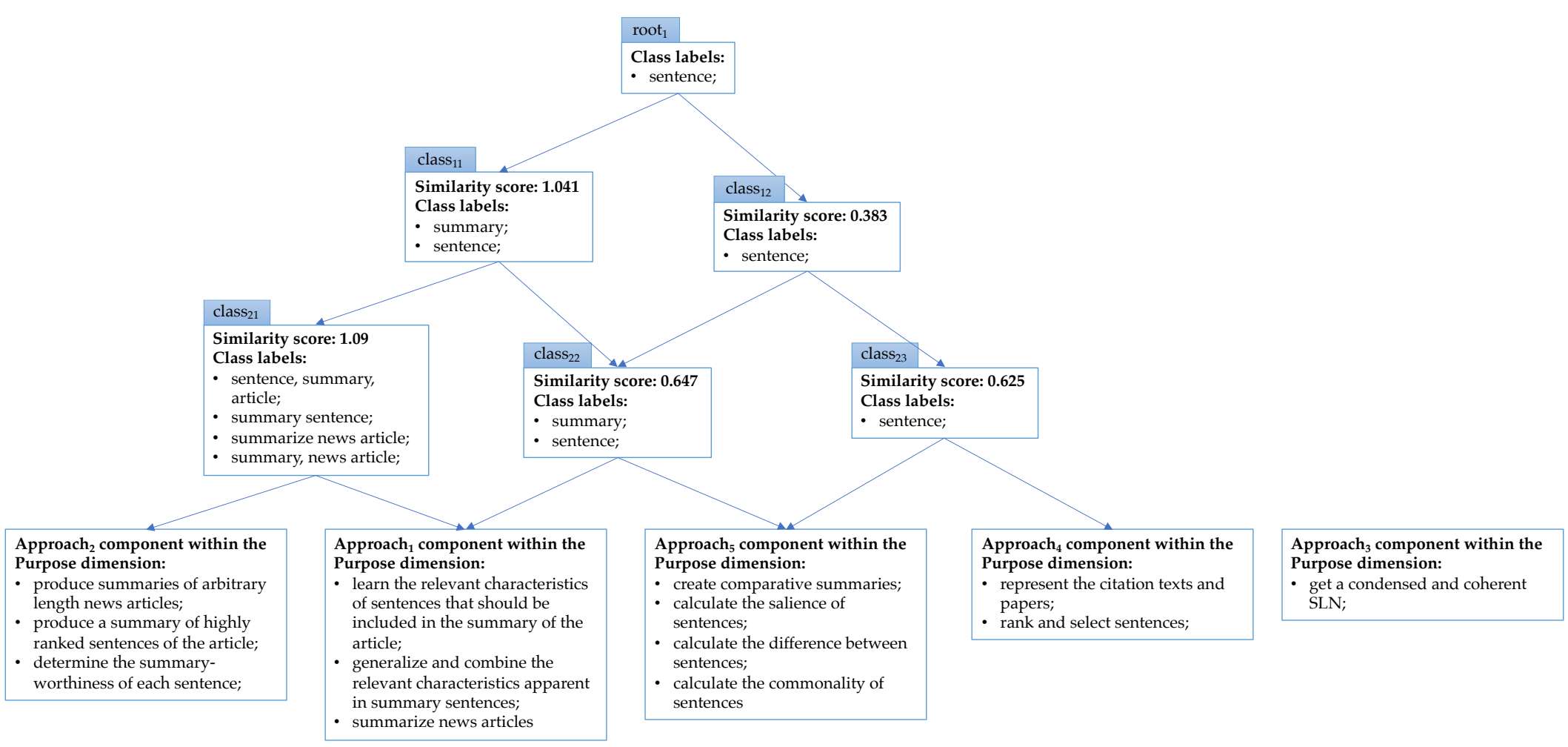

Figure 7. An example of class trees on the Purpose dimension

## 3.3 Comparison and Analysis

### *3.3.1 Comparison with Different Similarity Measures*

A tree-structure-based similarity measure is proposed to calculate the similarity between step components, where each step component is a short text. We compare this measure with existing text similarity measures.

Table 7 compares three methods for calculating the similarity between step components. Four step components extracted from scientific papers are as follows: a1: "neural network is trained", a2: "train neural network", a3: "neural network is trained on a corpus of articles", and a4: "neural network is modified". These four step components all related to the operation of neural networks. Semantically, a1 and a2 are identical, and they are more similar to a3 than to a4. This is because the operations in a1, a2, and a3 involve "train" neural network, while the operation in a4 involves "modify" neural network. The common-word-based similarity measure represents each text as a collection of words or phrases and calculates the similarity between texts as the sum of the weights of common words shared among these texts [10]. The SBERT model generates embedding vectors for the input text sequences and then calculates the cosine similarity between these embedding vectors [25].

Table 7. Comparison of three measures for calculating the similarity between two step components and their resulting class labels.

| Similarity/ Label | tree-structure-based | common-word-based (weight=1) | SBERT |
|---|---|---|---|
| **(a1, a2)** | 1 | 6 | 0.791 |
| | "train neural network" | "train"; "neural network" | - |
| **(a1, a3)** | 0.83 | 6 | 0.599 |
| | "train neural network" | "neural network"; "train" | - |
| **(a1, a4)** | 0.5 | 4 | 0.732 |
| | "neural network" | "neural network" | - |
| **Rank** | (a1, a2) > (a1, a3) > (a1, a4) | (a1, a2) = (a1, a3) > (a1, a4) | (a1, a2) > (a1, a4) > (a1, a3) |

Table 7 illustrates that only the tree-structure-based similarity measure achieves a reasonable similarity ranking. This is because it focuses on both common semantics and syntactic structure, aligning with the characteristics of human comparison of sentences. The common-word-based similarity measure calculates the same similarity score for (a1, a2) and (a1, a3), which is higher than the score for (a1, a4). This is because the first two pairs contain the identical common words "train" and "neural network", while the third pair contains only "neural network" as a common word. The SBERT model calculates a higher similarity score for (a1, a4) than for (a1, a3), which does not align with the similarity degree of actual semantics expressed by these three sentences. Furthermore, both the tree-structure-based similarity measure and the common-word-based similarity measure generate class lables that interpret the obtained similarity scores. The tree-structure-based measure captures more complete common semantics compared to the common-word-based measure.

Table 8 compares three measures in terms of their ability to distinguish pairs of sentences with varying degrees of similarity. Specifically, a1, a2, a3, and a4 are the same as in Table 7, while b1: "perform our query expansion method and update the query" and b2: "decompose documents into segments and retrieve those which are most relevant to query Q" are step components extracted from different approach components in scientific papers. This section uses the distinguish degree, similar to [10], to compare the ability of the three measures to distinguish different classes of step components. The distinguish degree is computed as: distinguish degree = (similarity degree of more similar pair - similarity degree of more dissimilar pair) / (similarity degree of more similar pair + similarity degree of more dissimilar pair). Semantically, the distinguish degree between (a1, a2) and (b1, b2) is higher than that between (a1, a3) and (b1, b2), and both are higher than the distinguish degree between (a1, a4) and (b1, b2). Because the verb and its object are identical in a1, a2, and a3, but a3 contains additional information expressed by adverbial, while a1 and a4 have different verbs but the same object, and b1 and b2 have different verbs and different objects.

Table 8. Comparison of three measures for distinguishing pairs of sentences with varying degrees of similarity

| Similarity | tree-structure-based | common-word-based (weight=1) | SBERT |
|---|---|---|---|
| **(a1, a2)** | 1 | 6 | 0.791 |
| **(a1, a3)** | 0.83 | 6 | 0.599 |
| **(a1, a4)** | 0.5 | 4 | 0.732 |
| **(b1, b2)** | 0.171 | 3 | 0.32 |
| **distinguish degree: (a1, a2); (b1, b2)** | 0.708 | 0.333 | 0.424 |
| **distinguish degree: (a1, a3); (b1, b2)** | 0.658 | 0.333 | 0.303 |
| **distinguish degree: (a1, a4); (b1, b2)** | 0.49 | 0.143 | 0.392 |

The results given in Table 8 show that the tree-structure-based similarity measure has a higher distinguish degree than other measures for distinguishing text pairs with varying degrees of similarity. Furthermore, only the tree-structure-based similarity measure achieves correct distinguish degree ranking for different combinations of comparisons. Because the tree-structure-based measure is suitable for comparing steps and sentences, where the similarity of the major syntactic constituents plays a significant role in the overall similarity. The common-word-based measure obtains the same distinguish degree for the first two comparisons in Table 8 because it focuses on the common semantics between texts, and (a1, a2) and (a1, a3) contain identical common words. This measure is suitable for computing the similarity of multiple texts containing multiple sentences, focusing on common semantics among texts rather than between individual sentences. The SBERT model calculates a higher distinguish degree between (a1, a4) and (b1, b2) than between (a1, a3) and (b1, b2), as it calculates a higher similarity score for (a1, a4) than for (a1, a3).

### *3.3.2 Comparison with different clustering algorithms*

We compare the proposed clustering algorithm and existing hierarchical clustering algorithms, including the Cluster-Expansion-on-Common-Words (CECW) algorithm [10] and the agglomerative clustering algorithm [26]. The CECW is a bottom-up text clustering method based on common words, which contains two main steps: (1) generating 2-element classes by finding the most similar text for each text based on the common words between two texts, and (2) expanding the k-element classes by adding one most similar text based on the common words between the k-element class and the text. The agglomerative clustering algorithm merges the most similar cluster pair in each iteration, starting with each text as an individual cluster.

Table 9 compares the efficiency and structure of the three algorithms for constructing class trees. For n input texts, in the worst case, the input texts belong to the same class. The maximum height of the class tree constructed by the proposed clustering algorithm is n, where each leaf node is a single text (i.e., 1-element class) and the root node contains all the input texts (i.e., n-element class). The maximum height of the class tree constructed by the CECW algorithm is n-1, the leaf nodes are 2-element classes and the root node contains all the input texts. The maximum height of the class tree constructed by the agglomerative clustering algorithm is n, each leaf node is a single text and the root node contains all the input texts.

The time complexity of the proposed clustering algorithm is $O(n^3)$. In each iteration, the algorithm merges each text or class with its most similar text or class. Specifically, in the first iteration, the algorithm finds the most similar text for each text, the time complexity is $O(n*(n-1))$, i.e., $O(n^2)$. The worst case is that all generated n classes are retained, meaning no two texts satisfy the condition where one text is the most similar to the other, and vice versa. In subsequent iterations, the algorithm finds the most similar class for each class, with a maximum time complexity of $O(n*(n-1))$, i.e., $O(n^2)$. As the maximum height of the class tree is n, the maximum time complexity of the clustering process is $O(n^2*n)$, i.e., $O(n^3)$. The time complexity of the CECW algorithm is also $O(n^3)$. In the step of generating 2-element classes, the time complexity is $O(n*(n-1))$, i.e., $O(n^2)$. When expanding k-element classes, the maximum time complexity is $O(n*(n-k))$, i.e., $O(n^2)$. As the maximum height of the class tree generated by CECW is n-1, the maximum time complexity of CECW is $O(n^2*(n-1))$, i.e., $O(n^3)$. The time complexity of the agglomerative clustering algorithm is also $O(n^3)$ because the time complexity of each step is $O((n-k)*(n-k-1))$, i.e., $O(n^2)$, and the maximum height of the generated class tree is n.

Table 9. Comparison between three hierarchical clustering algorithms

| | Our algorithm | CECW | Agglomerative clustering |
|---|---|---|---|
| **Time complexity** | $O(n^3)$ | $O(n^3)$ | $O(n^3)$ |
| **Maximum height** | *n* | *n-1* | *n* |

Furthermore, for clustering approaches, where each approach comprises a collection of steps, the proposed clustering algorithm is suitable. In class trees, the class labels of each class reflect common semantics of the step components within the approach components in the class. During the clustering process, the comparison between approaches involves comparing their constituent steps, and the comparison between classes involves comparing their class labels. Generally, the class labels of a class containing a larger number of approaches tend to express more general information, whereas the class labels of a class with fewer approaches convey more specific information. Comparing classes with similar granularity facilitates the extraction of richer common information from both classes, thereby avoiding the significant loss of information when a smaller class is merged with a larger one. For example, the comparison of two approaches in Figure 4 generates the common information including "train neural network", "modify neural network", "neural network", and "network". However, if one of the approaches, such as $AP_1$, is compared with a class that has only "network" as its class label, the common information between $AP_1$ and the class is limited to "network", which results in the loss of significant specific information from $AP_1$. The class trees constructed by the proposed clustering algorithm are more suitable for approach querying. In contrast, both the CECW algorithm and the agglomerative clustering algorithm involve comparisons between class and single text, which is suitable for clustering texts but not for approaches.

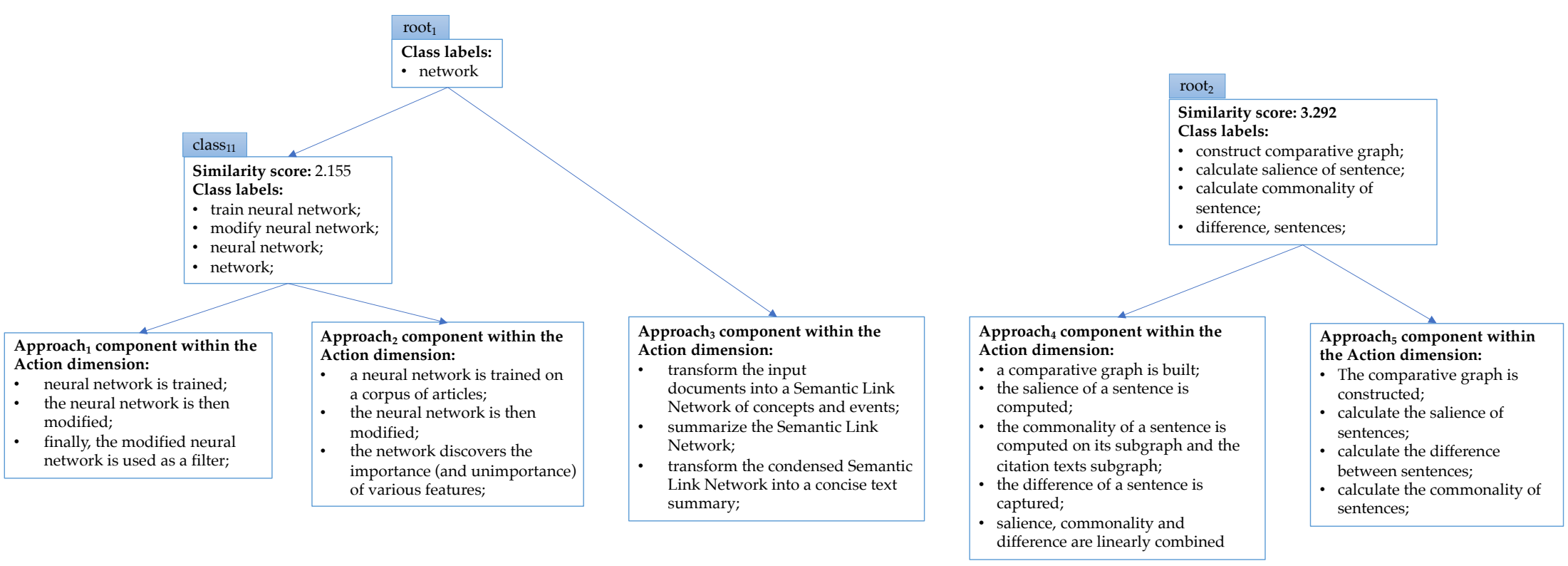

Figure 8. An example of class trees on the Action dimension after a pruning operation

Additionally, each step of both the proposed clustering algorithm and the CECW algorithm allows an approach or text to be classified into multiple classes. This aligns with the characteristics of approaches or texts, as an approach or text may share different common information with approaches or texts of different classes.

## 4 Approach Query on Multiple Dimensions

The five classification dimensions—action dimension, purpose dimension, manner dimension, condition dimension, and effect dimension—together form a multi-dimensional approach space that supports multi-dimensional queries.

The multi-dimensional approach space supports two types of queries: exact approach query and fuzzy approach query.

When performing exact querying on the multi-dimensional approach space, user input needs to conform to the fixed paradigm: query = {Action: string_a, Purpose: string_p, Manner: string_m, Condition: string_c, Effect: string_e}, where each string can be a word, phrase, clause, sentence, or Null. During the querying process, the system performs queries on the corresponding dimensions where the input is not Null.

Fuzzy approach querying refers to the process in which the user inputs a natural language query, and the system automatically extracts the corresponding query components across different dimensions based on predefined approach patterns. If the user query explicitly contains identifiers for each dimension, the system is capable of accurately identifying the corresponding query components of different dimensions and executing an exact query. For example, if the user query is "to document summarization based on data reconstruction", according to the approach patterns, "data reconstruction" corresponds to the manner (technique) of the approach, and "document summarization" corresponds to the purpose of the approach. In this case, the system queries "data reconstruction" on the Manner dimension, and queries "document summarization" on the Purpose dimension. However, in cases where the user query lacks some identifiers, the system is unable to precisely determine the query components for each dimension, necessitating the execution of a fuzzy query. For instance, if the user query is "document summarization based on data reconstruction", the query component "data reconstruction" corresponds to the manner (technique) of the approach, while "document summarization" cannot be accurately interpreted. Thus, the system performs exact query "data reconstruction" on the Manner dimension, and performs fuzzy query "document summarization" on each of other dimensions.

When querying on a dimension, our model represents the input within this dimension as a tree structure, as illustrated in Section 3.2.1, and compares the input with class labels or step components on the class trees of the dimension. The output is a set of approaches that meet the user query.

Before querying on the multi-dimensional approach space, pruning operations are applied to the class trees of each dimension. The specific procedure is as follows: for each class tree on each dimension, traverse all nodes from top-down, if the class label of a parent node is identical to that of one of its child nodes, the child node is removed, and all of its child nodes are directly connected to the parent node. This operation simplifies the structure of the class tree by removing redundant nodes, thereby improving query efficiency. Figure 8 and Figure 9 show the classification dimensions on the action and purpose dimensions after the pruning operations, respectively. As shown in the figures, the adjusted class trees no longer contain redundant nodes.

The multi-dimensional approach space supports a class-based querying mechanism that finds relevant classes by matching user input queries with class labels. Specifically, the class-based querying mechanism traverses the classes in class trees from top to bottom and calculates a matching score between each class and the query:

**Step 1:** The class-based query method calculates the matching score of each root class with the query and retains all trees with non-zero matching scores.

The matching score of each class with the query is calculated as the highest similarity score between each class label of that class and the query, using the tree-structure-based similarity measure. During the querying process, the similarity-weight of each class label and the query is set to 1.

**Step 2:** For each retained class (denoted as $class_r$), the query method calculates the matching score of each of its subclasses with the query, and ranks these subclasses based on their respective matching scores. If the subclass with the highest ranking exhibits a matching score that surpasses its parent class ($class_r$), the parent class is deleted, and the subclass is retained. Conversely, if the parent class's matching score is not surpassed by any of its subclasses, the parent class ($class_r$) is retained, and the traversal of all its subtrees is terminated.

**Step 3:** Repeat step 2 until no subtree of any of the retained classes can be traversed, i.e., until no retained class has a subclass with a matching score that surpasses the score of the retained class.

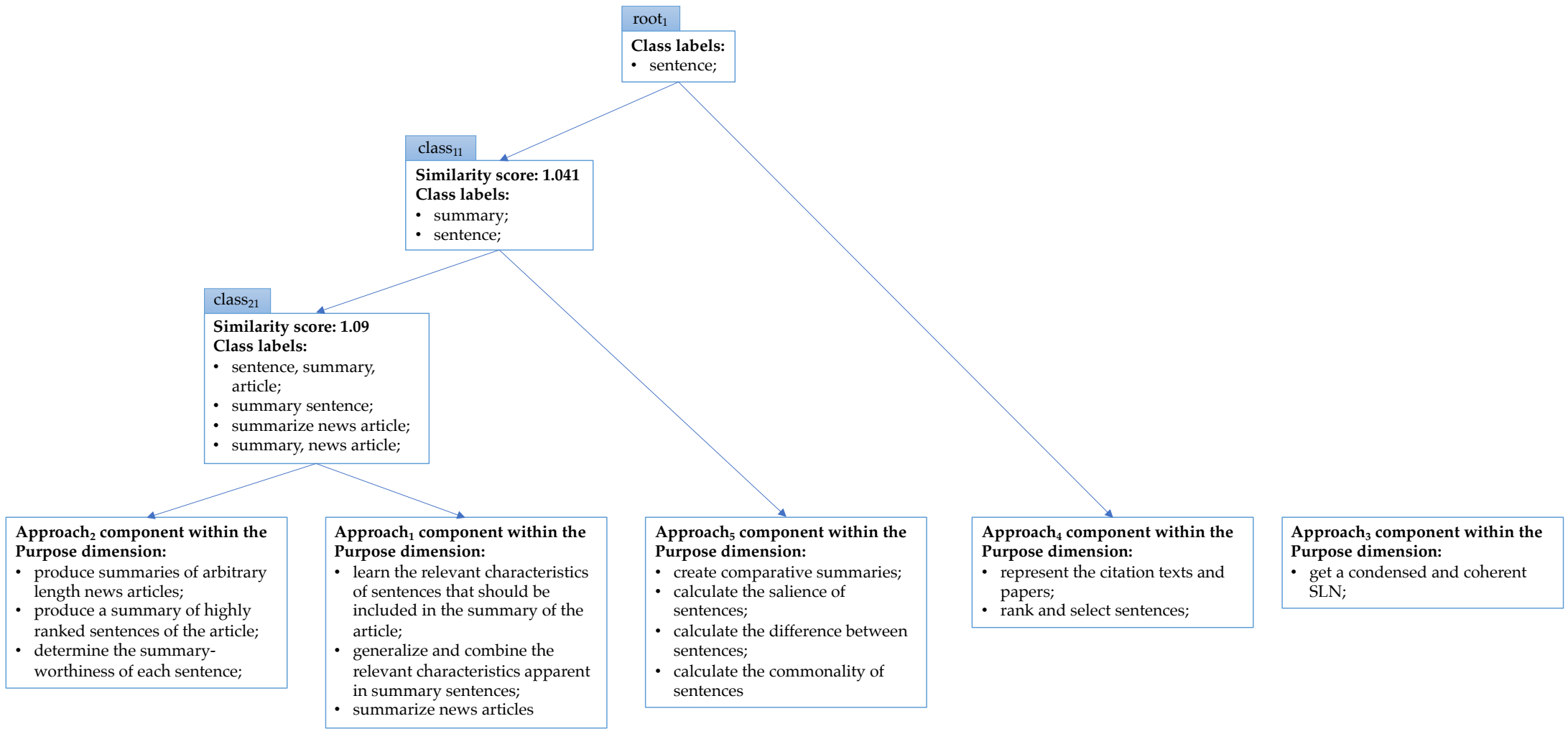

Figure 9. An example of class trees on the Purpose dimension after a pruning operation

**Step 4:** All retained classes are ranked according to their matching scores. Classes with class labels whose major syntactic constituents more closely match those of the query will be assigned higher ranking scores. If multiple classes have the same matching score, they are further ranked in descending order based on the weights of the class labels that matched the query.

**Step 5:** The approaches within the class are ranked according to the matching score of each approach component with the query.

The matching score of each approach component with the query is calculated as the highest similarity score between each step component within that approach component and the query, using the tree-structure-based similarity measure. During the querying process, the similarity-weight of each step component and the query is set to 1. Approaches whose major syntactic constituents of step components more closely match those of the query will receive higher ranking scores.

For example, if a user queries "neural network" on the action dimension shown in Figure 8, the class tree rooted at $root_1$ is retained because the matching score between this root node and the query is 0.75, which exceeds the threshold of 0. Next, the root node $root_1$ is replaced by its subclass $class_{11}$, since the matching score between $class_{11}$ and the query is 1, which is higher than 0.75. Finally, $class_{11}$ is retained because none of its subclasses have a matching score exceeding 1. Within $class_{11}$, $Approach_1$ and $Approach_2$ receive the same matching score of 0.75, so they are given the same ranking. The final output of the query includes $Approach_1$ and $Approach_2$.

It should be noted that if the matching scores between the query and the root nodes of all class trees on the current dimension are zero, this does not necessarily mean that there are no approaches on this dimension that satisfy the user's query. This is because, during the construction of the class trees, the merging of approaches or classes may lead to the loss of unique step elements or class label elements. For example, on the manner dimension shown in Figure 6, the step component "feature fusion" from $Approach_2$ is lost after merging with $Approach_1$. The merged class $root_1$ only retains the class label "neural network." To address this issue, one can traverse all the child nodes of the root nodes across all class trees on the current dimension and perform the operations described in Step 1. This process is equivalent to removing all root nodes, treating all subtrees under the root as new class trees, and then re-executing the query based on these newly formed trees. For instance, when querying "feature fusion" on the manner dimension of Figure 6, the matching scores between the query and the root nodes $root_1$ and $root_2$ of the two class trees are both zero. At this point, it becomes necessary to remove the root nodes $root_1$ and $root_2$, and directly traverse their child classes—namely, $Approach_1$, $Approach_2$, $Approach_4$, and $Approach_5$.

In summary, when querying on a dimension of the multi-dimensional approach space, the first step is to identify the target classes, and then rank the approaches within these classes. In most cases, it significantly reduces the search space and improves search speed compared to traditional keyword-based querying, as it does not require an exhaustive scan of all approaches. However, in certain special cases, for example when the query does not match any classes on the class tree, the class-based query may take longer than a keyword-based query. This is because, in such cases, the class-based query needs to traverse the entire class tree to confirm the absence of matches, whereas a keyword-based query might more quickly determine that no match exists.

The multi-dimensional approach space allows users to perform queries within specific dimensions based on their specific intent, which can effectively reduce irrelevant search results. It significantly enhances the capability of targeted searching, enabling users to more precisely locate the desired approaches and improving both retrieval efficiency and accuracy. For example, if a user inputs the query "summarize news articles" on the purpose dimension, indicating the intent to find approaches capable of summarizing news articles, then the approaches $Approach_1$ and $Approach_2$ would be returned. If the user inputs the query "feature fusion" on the manner dimension, aiming to find approaches that employ feature fusion techniques, then $Approach_2$ would be returned. If the user queries "construct graph" on the action dimension, seeking all approaches involving the construction of graphs, then $Approach_4$ and $Approach_5$ would be returned.

The multi-dimensional approach space allows users to perform queries across multiple dimensions simultaneously, meeting more complex query requirements. It supports locating classes from multiple dimensions at the same time, and the query result is the intersection of the approach classes returned by each dimension. For example, if a user intends to search for approaches that use feature fusion technique to address the problem of automatically generating summaries for news articles, they can input the query "summarize news articles" on the purpose dimension and "feature fusion" on the manner dimension. The model will search simultaneously in both the purpose and manner dimensions. The purpose dimension returns $Approach_1$ and $Approach_2$, while the manner dimension returns $Approach_2$. Ultimately, the model returns the intersection of the results from the two dimensions, that is, $Approach_2$.

In contrast, traditional keyword-based querying cannot accurately distinguish user intent, potentially leading to the retrieval of a large amount of irrelevant information. For example, when a user wants to find specific approaches aimed at training neural networks, a keyword-based query would return $Approach_1$ and $Approach_2$, even though they do not align with the user's actual needs. Because in these two approaches, "train neural network" appears as an action, rather than as a purpose. Therefore, such queries can generate many irrelevant results. Furthermore, traditional keyword-based querying approaches are typically limited to searching within a one-dimensional space. Users find it difficult to refine their queries across different dimensions, resulting in less precise or less relevant retrieval outcomes.

## 5. Related Works and Discussion

### 5.1 Similarity Calculation between Collections

An approach consists of a collection of steps, calculating the similarity between approaches entails computing the similarity between the respective collections of steps.

There exist many methods for calculating the similarity between collections. The common-word-based similarity measurement method represents each text as a collection of words and measures the similarity of a group of texts by calculating the sum of the weights of common words shared among those texts [10]. The Jaccard similarity [11][12] and Tanimoto coefficient [13][14] can also be used to calculate the similarity between collections. The Jaccard similarity is defined as the ratio of the size of the intersection to the size of the union of two collections. The Tanimoto coefficient, also known as the Jaccard index with augmentation, takes into account not only the presence or absence of members in collections but also their associated weights or values. In cases where members are binary (i.e., they exist or do not exist without further distinction), the Tanimoto coefficient is equivalent to the Jaccard index. While in cases where members are not binary but possess quantitative attributes, the Tanimoto coefficient exhibits similarities to cosine similarity in assessing the degree of similarity between two collections or vectors. In addition, the commonly used distance-based similarity measurements, such as Cosine distance and Euclidean distance [15][16], calculate the angular distance and point distance between two vectors in vector space. These metrics can also be applied to calculate the similarity between collections. Specifically, each text consists of a collection of words and can be represented as a vector within a word space, enabling the utilization of distance-based similarity measurements to quantify the similarity between text vectors.

These collection similarity calculation methods are not suitable for computing the similarity between approachs (step collections). Unlike the aforementioned collections where members consist exclusively of words or numerical values, in this paper, the members in collections are various linguistic units, ranging from individual words and phrases to clauses, sentences, and even multiple sentences. Therefore, the assessment of similarity between approaches (step collections) should not rely solely on simple common-word relationships, as word-level co-occurrence may overlook important syntactic-level similarities that exist between distinct steps. Distance-based similarity measurement methods are insufficient for measuring the similarity between step collections, as concatenating the steps in a collection into a vector representation would result in a substantial loss of semantic information. Furthermore, different approaches may contain one or multiple similar steps, emphasizing the necessity of comparing different members in different collections. This need cannot be adequately addressed by merely considering word co-occurrence, lexical similarity, or vector-based similarity measures.

### 5.2 Approach Extraction

Previous studies on approach extraction have primarily concentrated on the extraction of approach sentences or approach entities.

A system comprising two major steps was proposed for the identification of methodological sentences and the extraction and classification of methodological segments [3]. The first step involves the automatic identification of methodological sentences. Each sentence within scientific papers was represented as a feature vector, and then a classifier (such as SVM, decision trees, etc) is trained to classify sentences into one of the seven classes: Background (background knowledge), Aim (goal of research), Basis (specific other work that the presented approach is based on), Contrast (contrasting and comparison to other solutions), Other (specific other work), Textual (textual structure of the paper), and Own (own work including method, results, future work), which is same as the process suggested by [17]. The sentences within the Own category were further categoried into three subcategories: Solution (sentences that describe the methodology used in the paper), Result (sentences that contain the results presented in the paper), and Own_else (any other Own sentence that cannot be categorised as Solution or Result) using the same classifier. The second step involves the extraction and classification of methodological segments (phrases). Sentences that have been classified as belonging to the Solution category were then processed to identify and classifies methodological segments into four semantic categories, including Task, Method, Resource/Feature and Implementation. This process is formulated as a sequence tagging problem, and four separate phrase-based Conditional Random Fields (CRFs) with the lexical features (such as phrase type), syntactic features (such as dependence relation), semantic feature (such as the category of verbs), frequency feature, and others as features were trained to achieve this classification.

Another method uses structured abstracts as training data, where the sentences within the abstracts are divided into four distinct types: Purpose, Design/methodology/approach, Findings, and Other [4]. Methodological sentences are obtained from the Design/methodology/approach type, and non-methodological sentences are from the remaining types. Next, the classifiers, namely Convolutional Neural Network (CNN) and Bi-Directional Long Short-Term Memory (BiLSTM), are trained using pre-trained word embedding vectors as input to classify sentences into methodological sentences and non-methodological sentences. Finally, to extract methodological sentences more precisely, a rule-based method is applied, which utilizes words with low document frequency in different types of sentences to filter out sentences that may not be methodological sentences.

These methods extract approach entities (i.e., words or phrases) or single approach sentences, which convey limited information and often lack the necessary context required to fully comprehend the complex approaches employed. Additionally, existing methods for approach extraction typically rely on classifiers trained using manually annotated corpora. The performance of these classifiers is constrained by the quantity and quality of the training data, and they often fail to consider the linguistic characteristics of the approaches. Conversely, this paper focuses on extracting complete approaches that encompass a series of consecutive steps, and a single step typically contains one or multiple sentences. Our model extracts approaches based on approach patterns, which are identified according to five discourse relations capable of delineating steps or approaches. Furthermore, these patterns consider the occurrence of identifiers for each type of relations in various syntactic roles, facilitating a more comprehensive extraction of approaches from scientific papers.

### 5.3 Constructing Multi-dimensional Resource Space

The Resource Space Model is a multi-dimensional classification space for efficiently managing and categorizing resources from different dimensions [2][18][19]. Automatically constructing multiple classification dimensions from a given set of resources is a foundational problem of the research on the Resource Space Model (RSM) [20].

The techniques for constructing dimensions on texts include two categories: (1) extracting representations (e.g., words) from texts to represent dimensions on the texts, and (2) constructing class trees from texts so that each tree or a set of trees represent a dimension. A method to extracting candidate representations based on topic model, semantic graphs, semantic communities was proposed [22]. To make the sense of the dimensions, the keywords were mapped onto Wikipedia concepts by calculating the word relatedness using WordNet and constructing corresponding semantic graphs. Constructing the semantic link network on texts is an approach to representing and constructing class trees by making use of semantic relations between texts [23]. Another method that clusters texts based on common words of texts to construct class trees was proposed [24]. This method uses common words to represent texts and measures the similarity of a class of texts by calculating the sum of the weights of common words of the class. A bottom-up text clustering algorithm, Cluster-Expansion-on-Common-Words (CECW), is proposed to construct class trees of texts. The common words of each class on the class trees are used as the label of the class to indicate the common semantics of the class and manage the texts of the class. The class trees are regarded as a dimension on the texts. Different dimensions can be formed by assigning different weights to words within texts.

This paper aims at constructing a multi-dimensional approach space. Five dimensions for categorizing approaches are identified based on approach patterns, which are identified according to five discourse relations capable of delineating steps or approaches. Our model also clusters approach components within each dimension to construct the class trees for that dimension. Unlike previous common-word-based work, in this paper, each approach component is a collection of step components, where each step component can be various linguistic units, not just a single word. Our model uses a tree-structure-based similarity measure to compute the similarity between different step components within different approach components, and a collection similarity measure to compute the similarity between different approach components. Besides, our model uses a bottom-up clustering algorithm, which merges each approach component or class with its most similar approach component or class in each iteration, to construct class trees for the approach compoennts within each dimension. The class labels generated during the clustering process indicate the common semantics of the step components within the approach components in each class and are used to manage the approach components within the class.

## 6 CONCLUSION

This paper is dedicated to constructing a multi-dimensional approach space to manage approaches within scientific papers. The contribution consists of the following points:

- A method for identifying approach patterns from four linguistic levels in a top-down way is proposed, which gradually refines approach patterns from the semantic, discourse, syntactic, and lexical levels. The multi-level approach patterns are evaluated on the self-built dataset of scientific papers covering five different topics. Experimental results show that the precision of step extraction on the full-text dataset is 94.31%, with a recall of 93.64%. On the specific-section dataset, the precision improved by 1.68%. The advantage of the multi-level approach patterns in terms of extensibility is validated.

- Based on the approach patterns, five dimensions of approaches are derived: action, purpose, manner, condition, and effect. Experimental results show that each dimension achieves high coverage over the approaches, with the action dimension reaching 100%, and the other four dimensions all exceeding 70%. The high coverage indicates that the corresponding dimensions are capable of managing a larger number of approaches.
- A tree-structure-based similarity measure is proposed to calculate the similarity between steps (step components), focusing on syntactic-level similarities. This measure ensures that steps with similar major syntactic constituents attain a higher similarity score compared to those with similar minor syntactic constituents. Furthermore, a collection similarity measure is proposed to compute the similarity between approaches, as each approach is a collection of steps. The combination of these two similarity measures results in approach components with step components that share similar major syntactic constituents receiving a higher similarity score compared to those with step components sharing similar minor syntactic constituents.
- A bottom-up clustering algorithm is proposed to construct class trees for approach components within each dimension by merging each approach component or class with its most similar approach component or class in each iteration based on the proposed similarity measures. The clustering process allows an approach component to belong to multiple classes, as it may share different common step components with approach components of different classes. Experimental results confirmed that approaches classified into the same class on some dimensions may not belong to the same class on other dimensions, and may even exhibit significant differences.
- An application of approach queries on the constructed multi-dimensional approach space is discussed, and a class-based query mechanism is described. The multi-dimensional approach space supports queries within specific dimension as well as multi-dimensional queries, ensuring strong relevance between user queries and results. The class-based query mechanism rapidly reduces the search space and enhances search speed. It assigns higher ranking scores to classes or approaches whose class labels or steps exhibit greater relevance to the query in terms of major syntactic constituents.

## ACKNOWLEDGMENT

This work was supported by the National Science Foundation of China (project no. 61876048 and 61640212).

*Professor Hai Zhuge is the corresponding author

## APPENDIX

## A PATTERN OF GENERAL DESCRIPTION

The general description summarizes and generalizes the overall process of an approach. Typically, it does not enumerate specific steps but rather provides a concise overview of the fundamental information about the approach, including who proposed the approach, the number of steps the approach comprises, and a one-sentence summary of the overall content of the approach.

In addition, there are other types of general descriptions in which relations exist between the general descriptions and their respective steps. One such relation is that the general description outlines the overall purpose of the subsequent steps. Another is that the general description summarizes the result or effect achieved by executing those steps.

Generally, the pattern of general description can be generalized as follows:

```
<general description> ::= <general description
    about fundamental information> | <general
    description with purpose relation> | <gen-
    eral description with effect relation>
```

, where the <general description about fundamental information> contains three subtypes, i.e.,

```
<general description about fundamental infor-
    mation> ::= <fundamental information of
    type one> | <fundamental information of
    type two> | <fundamental information of
    type three>
```

The lexical items that explicitly express the meaning of approach such as "approach", "method", "framework", "model", and similar terms, can help to identify the general description about fundamental information. The general description with purpose relation or effect relation can be identified by analyzing lexical items that explicitly indicate a purpose or effect. From a syntactic perspective, sentences that belong to different types of general descriptions adhere to different requirements regarding their syntactic roles, including subject, verb, object, and adverbial.

(1) General description about fundamental information regarding "who proposed what specific approach"

The first type of general description about fundamental information involves the meaning of "who proposed what specific approach". Within this type, the general description should conform to certain criteria regarding the subject, verb, and object. Our focus is on extracting the approaches proposed within each individual scientific paper itself. Thus, the subjects of active-voice sentences ought to be expressions like "we", "our model", "this paper", or other analogous terms.

By analyzing examples of general descriptions within this type, the syntax pattern of this type of general descriptions is as follows:

```
<fundamental information of type one> ::= [<ad-
    verbial contain identifier>] (<subject
    contain identifier> <verb contain identi-
    fier> <object> | <subject contain identi-
    fier> <verb> <object contain identifier> |
    <object> <verb-ed contain identifier> |
    <object contain identifier> <verb-ed>)
    [<adverbial>]
```

For an active-voice general description of an approach in a paper, the subject of the sentence should indicate that the approach is proposed by the paper itself, rather than having been proposed by others. Thus, the subject typically adheres to the following pattern:

```
<subject contain identifier> ::= "we" | ("the"
    | "this" | "our") [<pre-modifier>] (<iden-
    tifier about paper-noun> | <identifier
    about method-noun>)
```

The <verb contain identifier> is typically a verb or verb phrase with a meaning similar to "propose", such as "put forward", "design", "present", "introduce", among others. Using lexicon to specialize the pattern of the verb, the specialized patterns of the verb is as follows:

```
<verb contain identifier> ::= "propose" | <syn-
    onym of propose> | "introduce" | <synonym
```

```
of introduce> | “design” | <synonym of de-
sign> | “present” | <synonym of present>
```

The <object contain identifier> is typically a noun or noun phrase that denotes the method, such as “an approach”, “a model”, “a framework”, or other similar terms. Using lexicon to specialize the pattern of the object, the specialized patterns of the object is as follows:

```
<object contain identifier> ::= (“a” | “an” |
    “the following” | <numeral word>) [<pre-
    modifier>] <identifier about method-noun>
```

In addition, the general descriptions may also incorporate adverbial phrases that signify the proposal of an approach within the respective paper, such as “in this paper”, “in our study”, and similar expressions. The adverbial phrases are primarily in the form of prepositional phrases. Using lexicon and syntax to specialize the pattern of the adverbial in this type, the specialized pattern of the adverbial is as follows:

```
<adverbial contain identifier> ::= (“in” | “on”
    | “for”) (“the” | “this” | “our”) [<pre-
    modifier>] (<identifier about paper-noun>
    | <identifier about method-noun>)
```

(2) General description about fundamental information regarding “an approach contains <numeral> steps”

The second type of general description about fundamental information involves the meaning of “an approach contains <numeral> steps”. Within this type, the general description should also adhere to specific criteria regarding the subject, verb, and object. By analyzing examples of general descriptions within this type, the syntax pattern of this type of general descriptions is as follows:

```
<fundamental information of type two> ::= <sub-
    ject contain identifier> <verb contain
    identifier> <object contain identifier> |
    <object contain identifier> <verb-ed con-
    tain identifier> <subject contain identi-
    fier> | “there” <be> <object contain iden-
    tifier> (“in” | “within” | “on”) <subject
    contain identifier>
```

Within this type, the subjects should express a specific approach or refer to a specific approach. In one case, the subjects explicitly convey meanings akin to “our method”, employing phrases such as “the proposed model”, “our approach”, or similar expressions. In another case, the subject is composed of pronouns or demonstrative words that refer back to the terminology used to describe the method in previous sentences, such as “ it”, “which”, “that”, and so forth. Using lexicon and syntax to specialize the pattern of the subject in this type, the specialized pattern of the subject is as follows:

```
<subject contain identifier> ::= (“the” |
    “this” | “our”) [“proposed”] [<pre-modi-
    fier>] <identifier about method-noun> |
    “it” | “which” | “that”
```

The <verb contain identifier> is typically a verb or verb phrase with a meaning similar to “contain”, such as “consist of”, “involve”, “incorporate”, among others. Using lexicon to specialize the pattern of the verb, the specialized patterns of the verb is as follows:

```
<verb contain identifier> ::= “contain” | “con-
    sist of” | <be> “composed of” | <synonym
    of contain>
```

The <verb-ed contain identifier> not only contains the passive voice of the above verb, it also contains the active-voice verbs like “constitute”, “form”, “make up”, and so on.

```
<verb-ed contain identifier> ::= “constitute” |
    “make up” | <synonym of constitute> | <be>
    (“contained” | <synonym of contained>)
    (“by” | “in” | “within” | “on”)
```

The <object contain identifier> is typically a noun phrase that conveys meanings akin to “<numeral> steps”, employing phrases such as “three steps”, “multiple phases”, “the following steps”, or similar expressions. In another case, the object might not include the aforementioned words that explicitly convey a meaning similar to “multiple steps”; instead, it could comprise several coordinate phrases, with each phrase representing the short name of a distinct step. Using lexicon to specialize the pattern of the object, the specialized patterns of the object is as follows:

```
<object contain identifier> ::= (<numeral word>
    | “multiple” | <synonym of multiple> |
    “following”) [<pre-modifier>] <identifier
    about step-noun> | (<noun phrase> | <par-
    ticiple phrase>) {(“,” | “;”) (<noun
    phrase> | <participle phrase>)} (“,” | “;”
    | “and”) (<noun phrase> | <participle
    phrase>)
```

If the object is composed of multiple coordinate phrases, the phrases are separated by commas or semicolons.

(3) General description about fundamental information regarding a one-sentence summary of the approach

The third type of general description about fundamental information involves the meaning of “our approach is …”, which is typically a one-sentence introduction of the overall content of the approach. Within this type, the general description should adhere to specific criteria regarding the subject and verb. By analyzing examples of general descriptions within this type, the specialized pattern of this type of general descriptions is as follows:

```
<fundamental information of type three> ::=
    <subject contain identifier> <be> [“that”
    | “to”] (<subordinate clause> | <verb
    phrase> | <participle phrase> | <noun
    phrase>)
```

The <subject contain identifier> is typically a noun phrase with a meaning similar to “our method”, such as “our method”, “the proposed model”, “our approach”, among others. The specialized pattern of the subject is as follows:

```
<subject contain identifier> ::= (“the” |
    “this” | “our”) [“proposed”] [<pre-modi-
    fier>] <identifier about method-noun>
```

(4) General description expressing purpose

If the general description of an approach describes the purpose of its subsequent steps, the general description may be a sentence, an adverbial phrase, or an adverbial clause that express the purpose. The syntax pattern of this type of general descriptions is as follows:

```
<general description with purpose relation> ::=
      <adverbial contain purpose identifier> |
      <subject> <verb contain purpose identi-
      fier> <object> | <subject contain purpose
      identifier> <verb> <object>
```

, where the <adverbial contain purpose identifier>, <verb contain purpose identifier>, and <subject contain purpose identifier> are the same as those presented in *section 2.1.2*.

(5) General description expressing effect

If the general description of an approach summarizes the result or effect achieved upon executing its corresponding steps, the general description is typically a sentence that express the effect, in which the syntactic roles identifying the effect may be the adverbial or the verb of the sentence. The syntax pattern of this type of general descriptions is as follows:

```
<general description with effect relation> ::=
      <adverbial contain effect identifier>
      <subject> <verb> [<object>] | <subject>
      <verb contain effect identifier> <object>
```

, where the <adverbial contain effect identifier> and <verb contain effect identifier> are the same as those presented in *section 2.1.4*.

The patterns are confirmed by the analysis of general descriptions identified in scientific papers' approaches.

## B Extend Approach Patterns to Pseudocode

In some scientific papers, authors may use pseudocode to describe their approaches. We extend the approach pattern to enable the processing of pseudocode for various approaches as well.

Just like a step can include one or multiple sentences, in pseudocode, each step comprises one or multiple program statements.

The basic statements in programming code include assignment statements, expression statements, declaration statements, input/output statements, conditional statements, looping statements, and jump statements. Among these, assignment, expression, declaration, input/output, and jump statements can each be regarded as a single step in the execution of a program. We refer to these five types of statements as executable statements. For conditional statement, one condition within the conditional structure is typically paired with a subsequent executable statement to constitute a step, where the executable statement represents the action to be performed, and the condition determines when that action should be executed.

Looping statements can be categorized into two cases. In the case where the number of iterations $l$ is predetermined, the statements contained within the body of the looping structure are executed $l$ times, an example of which is the For-loop. Another case arises when the number of iterations is not predetermined. In such a scenario, it is essential to evaluate the conditions after each iteration to determine whether the loop should proceed to the next iteration, a prime example of which is the While-loop. For each statement within the body of the looping structure, if the statement belongs to executable statements, i.e., excluding conditional or looping statements, each individual execution of that statement is regarded as a step. If the statement is a conditional statement, one condition is paired with a subsequent executable statement to constitute a step, i.e., the condition in conditional structure must be combined with other executable statement to form a step.

Table H.1 gives illustrative examples of the different types of statements and their corresponding steps. Compared to the steps outlined in natural language, each of these basic statement types contains an action or manner element. The assignment, expression, declaration, and input/output statements also contain the effect element, because these statements produce certain results or outcomes. The conditional statement contains the action, manner, condition, and effect elements, as it denotes the execution of executable statements under certain condition. The looping statement with a predetermined number of iterations contains the action, manner, and effect elements. While the looping statement with an unknown number of iterations contains the action, manner, condition, and effect elements, as it necessitates evaluating conditions after each iteration to determine whether the loop should proceed to the next iteration. The purpose element of pseudocode usually appears in annotations, which can be identified through the purpose identifier in natural language.

Table H.1. Different types of statements and their corresponding steps. Assume that the current step is *step i-1*.

| **Statement Type** | | **Example** | **Step** | **Five Dimensions** |
|---|---|---|---|---|
| assignment statement | | x=10 | step i: x=10. | action/manner; effect; |
| expression statement | | y=z+2 | step i: y=z+2. | action/manner; effect; |
| declaration statement | | int x | step i: int x. | action/manner; effect; |
| input/output statement | | return y | step i: return y. | action/manner; effect; |
| jump statement | | break | step i: go to *step k*. | action/manner; |
| conditional statement | | if condition 1: statement 1 | step i: if condition 1, statement 1. | action/manner; condition; effect; |
| looping statement | number of iterations is predetermined | for k∈M: statement 1 statement 2 | step i: statement 1. step i+1: statement 2. step i+2: statement 1. | action/manner; effect; |

<table>
<tr><td></td><td></td><td>//|M|=2</td><td>step i+3: statement 2.</td><td></td></tr>
<tr><td></td><td>number</td><td>while k < p do:</td><td>step i: if k<p, statement 1.</td><td>action/manner;</td></tr>
<tr><td></td><td>of iterations is unknown</td><td>statement 1<br>statement 2</td><td>step i+1: if k<p, go to step i; else, statement 2.</td><td>condition; effect;</td></tr>
</table>